\documentclass[lettersize,journal]{IEEEtran}
\usepackage{amsmath,amsfonts}
\usepackage{algorithmic}
\usepackage{algorithm}
\usepackage{array}
\usepackage{textcomp}
\usepackage{stfloats}
\usepackage{url}
\usepackage{verbatim}
\usepackage{paralist}
\usepackage{graphicx}
\usepackage{cite}
\usepackage{amsthm, amssymb}
\usepackage{amsmath}
\usepackage{caption}
\usepackage{upgreek}
\usepackage{longtable}
\usepackage{multirow}
\usepackage{subcaption}
\usepackage{tabularx}
\usepackage{bm}
\usepackage{gensymb}
\usepackage{fixltx2e}
\usepackage{tikz}
\usetikzlibrary{3d}
\usepackage{enumitem}
\usepackage{booktabs}
\usepackage{adjustbox}
\usepackage{subcaption}

\begin{document}

\title{A Coordinated Dual-Arm Framework for Delicate Snap-Fit Assemblies}
\author{Shreyas Kumar$^{1,*}$, Barat S$^{1}$, Debojit Das$^{1}$, Yug Desai$^{1}$, Siddhi Jain$^{2}$, Rajesh Kumar$^{2}$  \\ and Harish J. Palanthandalam-Madapusi$^{1}$ 
\thanks{The authors gratefully acknowledge support for this research from Addverb Technologies Pvt. Ltd.}
\thanks{$^1$ IITGN Robotics Laboratory, Department of Mechanical Engineering, Indian Institute of Technology Gandhinagar, Gujarat 382355, India.}
\thanks{$^2$ Addverb Technologies Limited, Noida, Uttar Pradesh 201301, India.}
\thanks{$^*$Corresponding author: {\tt\small shreyas.kumar@icloud.com}.}
}

\maketitle

\begin{abstract}
Delicate snap-fit assemblies, such as inserting a lens into an eye-wear frame or during electronics assembly, demand timely engagement detection and rapid force attenuation to prevent overshoot-induced component damage or assembly failure. We address these challenges with two key contributions. First, we introduce SnapNet, a lightweight neural network that detects snap-fit engagement from joint-velocity transients in real time, showing that reliable detection can be achieved using proprioceptive signals without external sensors. Second, we present a dynamical-systems-based dual-arm coordination framework that integrates SnapNet driven detection with an event-triggered impedance modulation, enabling accurate alignment and compliant insertion during delicate snap-fit assemblies. Experiments across diverse geometries on a heterogeneous bimanual platform demonstrate high detection accuracy (over 96\% recall) and up to a 30\% reduction in peak impact forces compared to standard impedance control.
\end{abstract}

\def\abstractname{Note to Practitioners}
\begin{abstract}
Automating snap-fit assembly is difficult due to tight alignment requirements and the brief force surge at engagement. We present a practical method where engagement is detected from joint-velocity transients using SnapNet. This detection initiates a rapid ramp-down of Cartesian stiffness along the insertion axis while preserving lateral stiffness for alignment. The method requires no external sensors; in our setup, inference ran at 100\,Hz with under 50\,ms latency. For deployment, we recommend restricting detection to the final approach, using low to moderate speeds (up to 10 mm/s), and tuning stiffness decay to part tolerances. The approach is effective when velocity transients are clearly observable; on robots with high friction or high gear ratios, joint-velocity signals may be insufficient and additional sensing may be needed. On platforms without those limitations, we observed lower peak contact forces and more reliable insertions, demonstrating the method’s suitability for practical snap-fit automation.
\end{abstract}


\section{INTRODUCTION}
\label{sec:introduction}
\begin{figure}[t]
    \centering
    \includegraphics[width=0.9\linewidth]{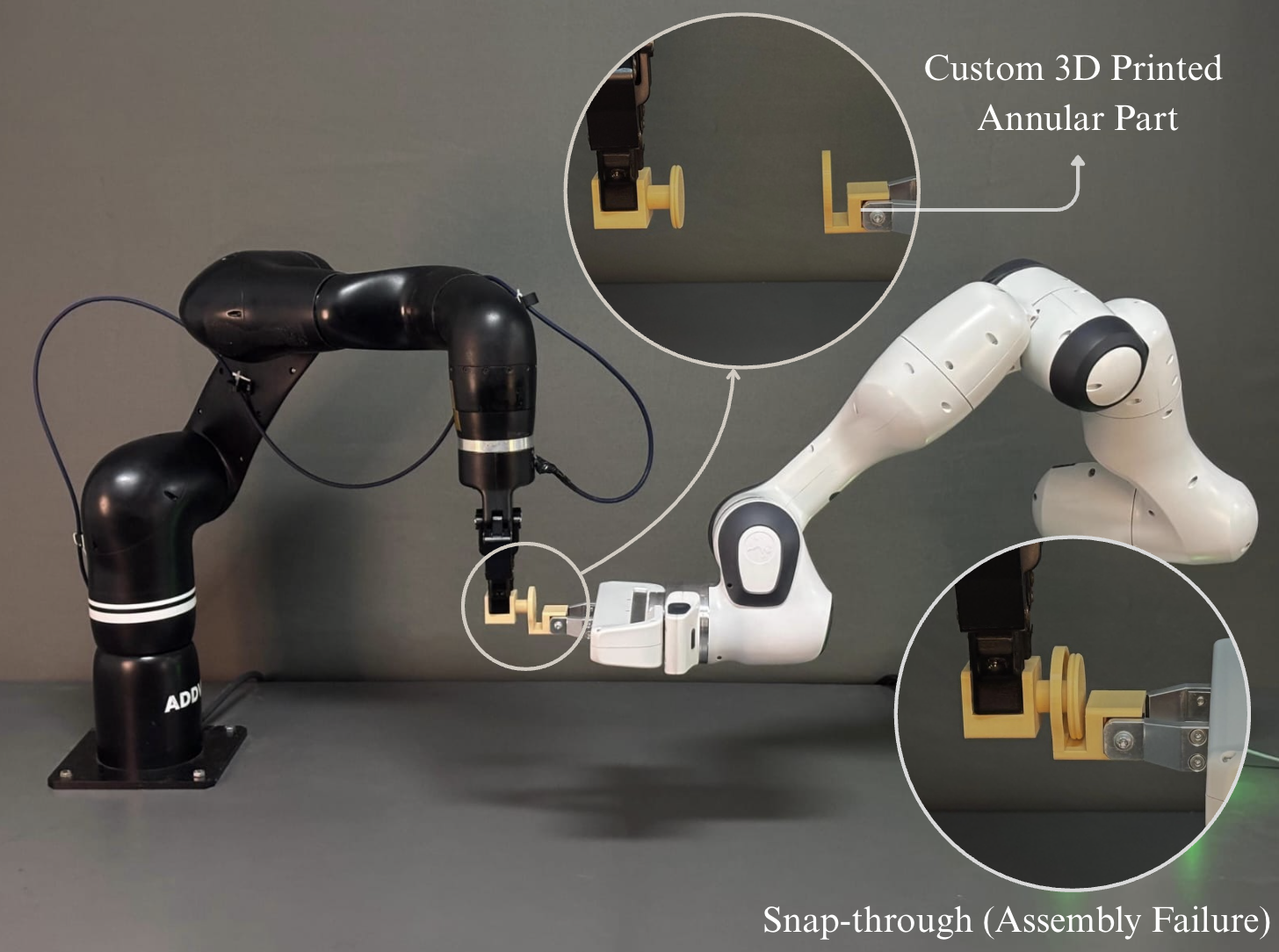}
    \caption{Bimanual snap-fit task performed under a unified framework using Franka FR3 and Addverb Heal Cobot. The proposed method is directly applicable to snap-fit assembly skills for bimanual/humanoid robots.}
    \label{fig:dual}
\end{figure}

\IEEEPARstart{S}{nap-fit} mechanisms are widely used in industrial assembly for their speed and cost-effectiveness. Automating these insertions with bimanual robots is especially difficult for delicate assemblies such as eyewear or electronic components, where tight tolerances, brief but forceful engagement events, and overshoot can easily cause damage or assembly failure (Fig.~\ref{fig:dual}).

\begin{figure}[t]
    \centering
    \includegraphics[width=0.9\linewidth]{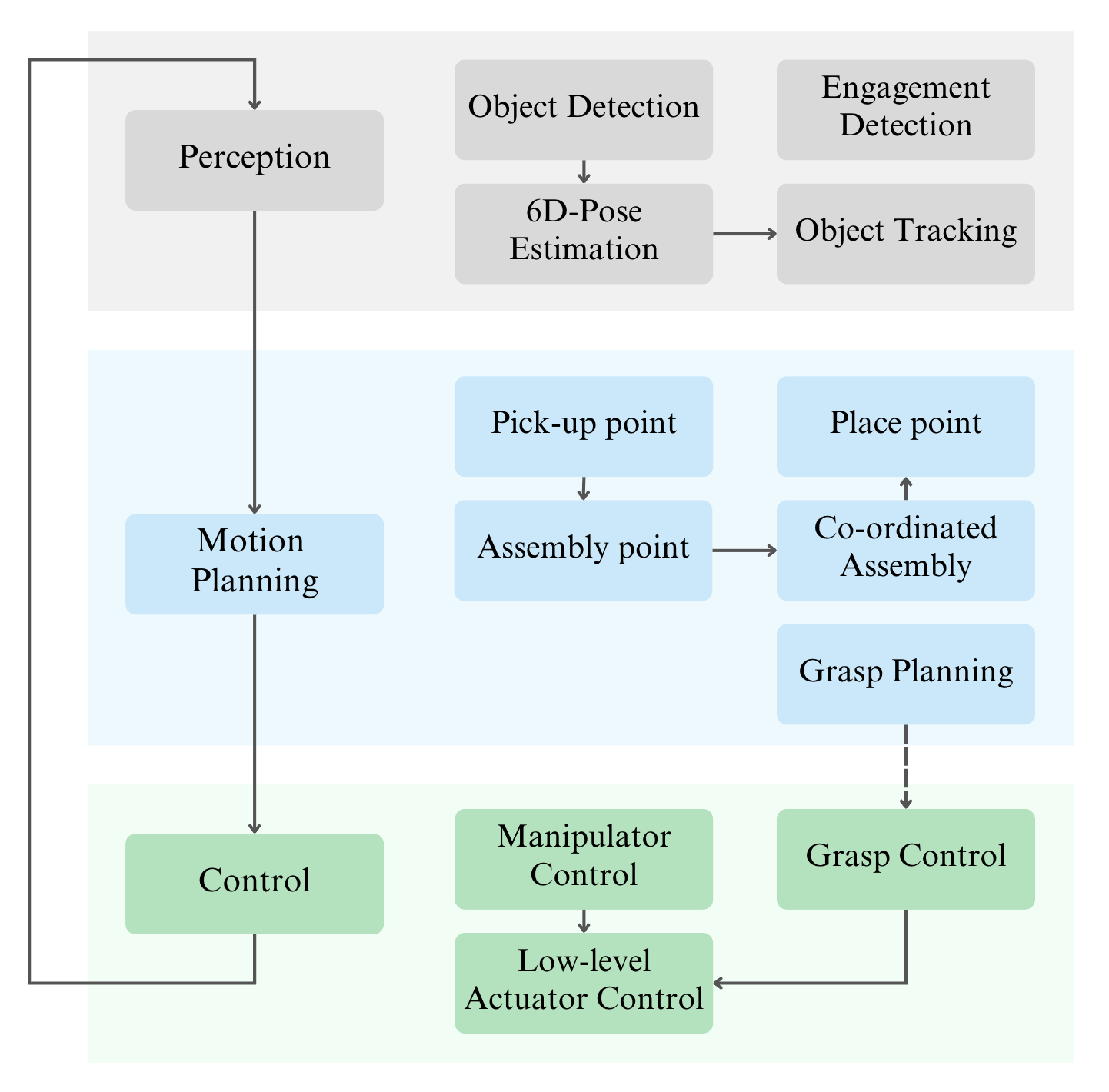} 
    \caption{Representative perception–planning–control pipeline for automated snap-fit assembly.  This study targets the shaded blocks: high-level motion planning, snap-fit engagement detection, and bimanual coordinated insertion.  All other modules are treated as given.}
    \label{fig:pipeline_overview}
\end{figure}

Despite the ubiquity of snap-fits, most assemblies, even for standard parts, remain manually executed on factory lines. As industry moves toward deploying humanoid or dual-arm manipulators for routine tasks, prior works have rarely addressed the full bimanual challenge associated with delicate snap-fit assemblies. Our work aims to narrow this gap. However, achieving a fully automated snap-fit assembly line is a multi-stage endeavor that spans perception, planning, and control (Fig.~\ref{fig:pipeline_overview}). A typical pipeline first localizes parts, estimates their poses, and tracks them during manipulation. High-level motion planners then select pick-up, placement, and assembly poses, while grasp planners assign feasible contacts. Finally, low-level controllers execute the planned trajectories and regulate the interaction forces during insertion.

This work concentrates on three stages critical to \emph{delicate} snap-fits: (i) high-level coupled motion planning to guide both arms from pick-up to insertion, (ii) snap-fit engagement detection, and (iii) an event-triggered impedance adaptation via variable impedance control (VIC), that limits the applied force during locking. We assume that the upstream perception and the downstream low-level actuation problems are solved.
\begin{table*}[b]
    \centering
    \begin{tabular}{@{} p{3.4cm} p{2.2cm} p{1.5cm} p{8.8cm} @{}}
        \toprule
        \textbf{Action} & \textbf{Force Sensitivity} & \textbf{Overshoot Propensity} & \textbf{Representative Tasks}  \\ 
        \midrule
        Push until barrier reached 
        & High & Low & 3\,C Electronics Assembly, \textbf{Type-C USB Cable}, Ethernet Port, 3.5\,mm Audio Jack  \\
        & Low  & Low & \textbf{Marker/Pen Caps}, Battery Covers, Snap-on Lids, Industrial Latches  \\
        \midrule
        Push without barrier 
        & High & High & \textbf{Lens insertion in Eyewear Frame}  \\
        & Low  & High & Window snap-fit fixtures \\
        \midrule
        Push/pull until release 
        & High & Moderate & Push Buttons, Memory Card Insertion  \\
        & Low  & Moderate & \textbf{Water bottle lid}  \\
        & Low  & Low & \textbf{Emergency-stop buttons}  \\
        \midrule
        Rotate until lock/release
        & High & Moderate & Mounting Camera Lens  \\
        & Low  & Low & Fan Regulator, Gas-stove Regulator  \\
        \bottomrule
    \end{tabular}
    \caption{Snap-fit taxonomy organized by action type, with corresponding force sensitivity (risk of damage) and overshoot propensity (propensity to move beyond nominal point of rest). For the purpose of this work, we term assembly tasks with high force sensitivity and high overshoot propensity as delicate assemblies.}
    \label{tab:taxonomy}
\end{table*}
Prior work on snap-fit detection, bimanual coordination, and compliant control has largely treated these components in isolation. For engagement detection, methods relied on force/torque transients and handcrafted features, often paired with classifiers to identify contact phases and failure modes \cite{doltsinis2019machine, kim2022robotic}. Deep learning has improved robustness to noisy signals through time-delay networks and recurrent models trained on raw or transformed force profiles, though often requiring retraining for new platforms or geometries \cite{sato2021force, cui2023fast}. Other sensing modalities such as acoustic \cite{rouwhorst2023acoustic}, tactile \cite{fang2025evetac}, and vision, offer precise cues but raise integration, cost, or occlusion challenges. Bimanual coordination approaches span role-based systems \cite{liu2023birp}, attention-based imitation learning \cite{motoda2025learning, lee2024interact}, and dynamical systems (DS) that enable smooth modulation between synchronized and decoupled motions \cite{figueroa2018unified, KhadivarAdaptiveFingers}. In compliant control, impedance and variable impedance control (VIC) have been extensively explored for regulating contact dynamics \cite{hogan1985impedance, buchli2009variable, abudakka2020review}. Reinforcement learning further adapts compliance across task contexts \cite{bogdanovic2020learningb, zhang2021learning, anand2024data}, while event-triggered control frameworks optimize gain transitions using force thresholds or state conditions \cite{mazo2010decentralized, girard2013dynamic}. Yet, many strategies assume external sensing and task-specific tuning.

Overall, existing methods either rely on additional sensing hardware or fail to integrate detection, coordination, and compliance within a unified framework. This paper addresses this gap through the following contributions:
\begin{itemize}
    \item \textbf{SnapNet for proprioceptive snap detection:} a lightweight neural network that identifies snap-fit engagement from joint-velocity transients in real time, demonstrating that reliable detection can be achieved using proprioceptive signals alone, without external sensors. Using only joint-velocity inputs also supports straightforward integration with standard robotic interfaces.
    \item \textbf{A unified framework for snap-fit assembly:} a dynamical-systems (DS) based dual-arm coordination scheme that synchronizes arm motions during transport, selectively decouples them during insertion, and integrates SnapNet driven detection with event-triggered impedance adaptation, thereby attenuating residual forces that would otherwise risk damage or assembly failure.
\end{itemize}

The remainder of this paper is organized as follows. In Section~\ref{subsec:mechanics}, we establish a task-oriented taxonomy of snap-fit assemblies and use it to design a benchmark spanning multiple geometries and fit types. From this benchmark, we compile a multimodal dataset that enables systematic comparison of sensing modalities, which motivates the development of SnapNet, introduced in Section~\ref{subsec:snapnet}. Building on this detection capability, we introduce a dynamical-systems (DS) based dual-arm coordination framework together with an event-triggered impedance adaptation (VIC) strategy in Section~\ref{subsec:ds} and \ref{subsec:vic}. Section~\ref{sec:experiments} presents the experimental setup and evaluates performance across control modes and geometries. Section~\ref{sec:discussion} discusses limitations, and Section~\ref{sec:conclusion} concludes with a summary and future directions.
\section{METHODOLOGY}
\label{sec:methodology}

\subsection{Snap-Fit Mechanics and Taxonomy}
\label{subsec:mechanics}
\subsubsection{Mechanics}
\label{subsubsec:mechanics}
Snap-fit engagement typically proceeds through elastic deformation, a rapid snap transition, and final locking. During deformation, a compliant feature stores strain energy; once a critical load is reached, set by geometry, undercut, and material, the feature releases this energy over a short interval and settles into the locked configuration \cite{gu2006evaluation, zhang2018plastic}. This rapid snap transition produces brief transients in contact force and on back-drivable arms, recognizable joint-level signatures.

\subsubsection{Task Taxonomy}
Snap-fit designs are commonly categorized as cantilever, torsional, or annular, depending on the elastic element and locking geometry. Despite these geometric differences, all exhibit a brief but distinctive transient at engagement~\cite{gu2006evaluation}, thus enabling proprioceptive detection in our framework. For controller design and evaluation, however, geometric classification alone is insufficient. We therefore organize tasks along two complementary axes more closely tied to sensing and control requirements.
\begin{itemize}
    \item \textbf{Force Sensitivity} captures how vulnerable the assembly is to residual forces. High-sensitivity tasks involve fragile or low-tolerance components (e.g., eyewear lenses), where excessive forces may cause fracture or misassembly. Low-sensitivity tasks involve robust or forgiving parts (e.g., pen caps, snap-on lids).  
    \item \textbf{Overshoot Propensity} reflects the likelihood that snap-through motion carries the part beyond its nominal resting point. Low-propensity tasks quickly settle against a barrier (e.g., pen caps, snap-on lids), while high-propensity tasks allow continued motion or rebound (e.g., eyewear lenses).  
\end{itemize}

This taxonomy is organized by action type, as shown in Table~\ref{tab:taxonomy}. The grouping highlights practical task families and provides representative benchmarks that span the range of \emph{compliance demands} relevant to snap-fit automation. In particular, tasks with either high force sensitivity and high overshoot propensity (termed as delicate assembly tasks) impose the strictest of requirements on post-engagement force regulation, \emph{motivating the use of accurate snap event detection and variable impedance control strategies} discussed in later sections.

\subsection{SnapNet}
\label{subsec:snapnet}

\subsubsection{Data Collection}
\label{subsec:data_collection}

To systematically study snap-fit signatures across sensing modalities, we designed our benchmark following the taxonomy in Table~\ref{tab:taxonomy}. The selected objects span the full range of \textbf{force sensitivity} and \textbf{overshoot propensity}, thereby capturing diverse mechanical behaviors—from low-sensitivity, low-propensity assemblies (e.g., pen caps, snap-on lids) to high-sensitivity, high-propensity cases (e.g., eyewear lenses). This coverage ensures that the resulting dataset encompasses both regimes where precise compliance adaptation is critical and those where standard impedance control suffices.

Approximately $500$ insertion trials were collected using a Franka Research $3$ (FR\,$3$) manipulator equipped with a Franka Hand gripper. The dataset includes household items (water bottle lid, marker/pen cap), consumer electronics (Type-C USB cable), industrial components (emergency-stop button), and delicate assemblies such as lens in an eyewear frame, along with custom 3D-printed geometries spanning varied locking profiles. Each trial consisted of a straight-line insertion under nominal alignment, executed at controlled speeds between $20$ and $100$\,mm/s, and continued until a clear engagement event or failure (e.g., misalignment, premature halt, or part fracture) occurred. Snap-fit engagement instances were \emph{manually annotated}, yielding a mix of successful and unsuccessful outcomes. All data streams were time-synchronized and sampled at $100$\,Hz across multiple sensing modalities.

\begin{figure}[h]
  \centering
  \begin{subfigure}{0.45\linewidth}
    \centering
    \includegraphics[width=\textwidth]{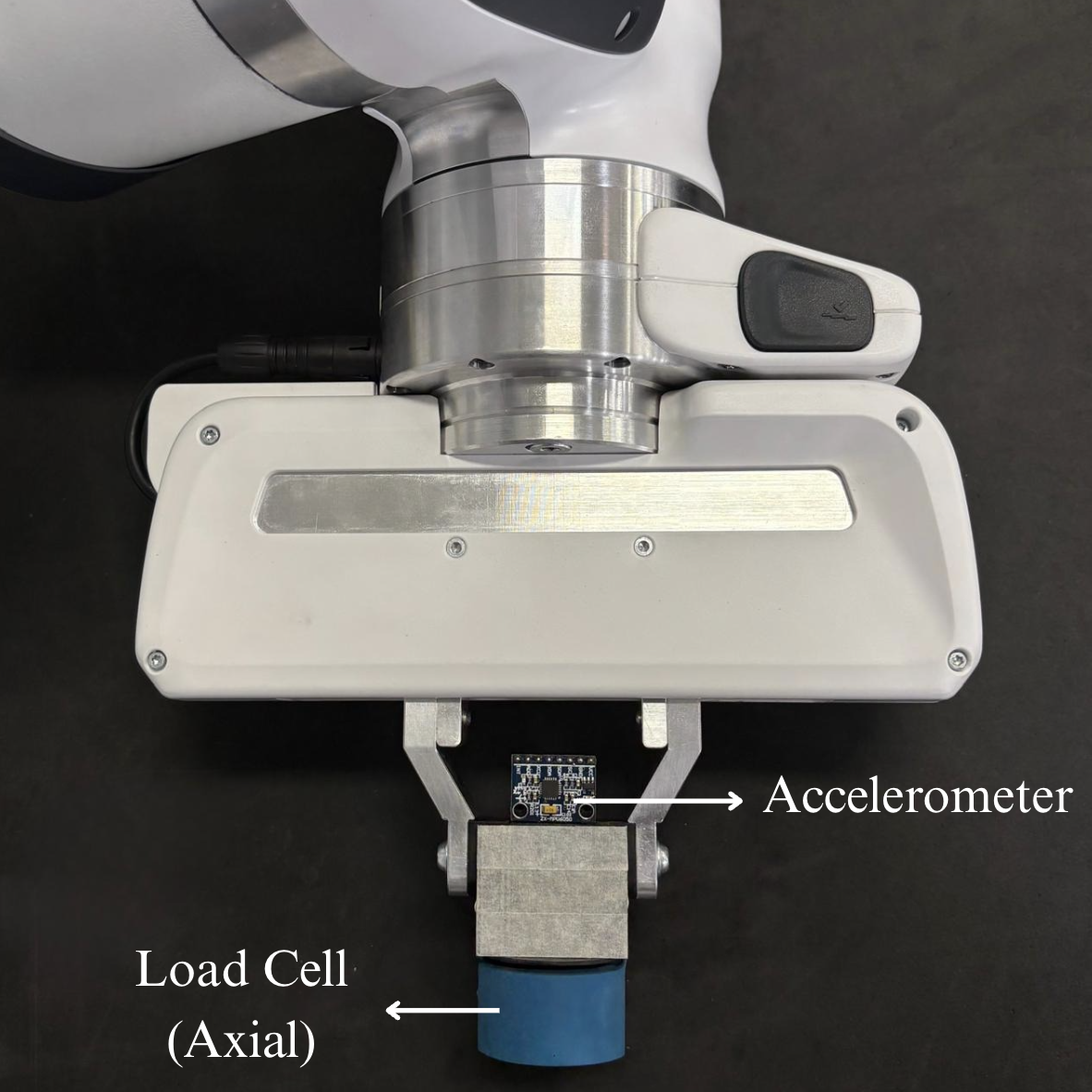}
    \caption{}
  \end{subfigure}
  \begin{subfigure}{0.45\linewidth}
    \centering
    \includegraphics[width=\textwidth]{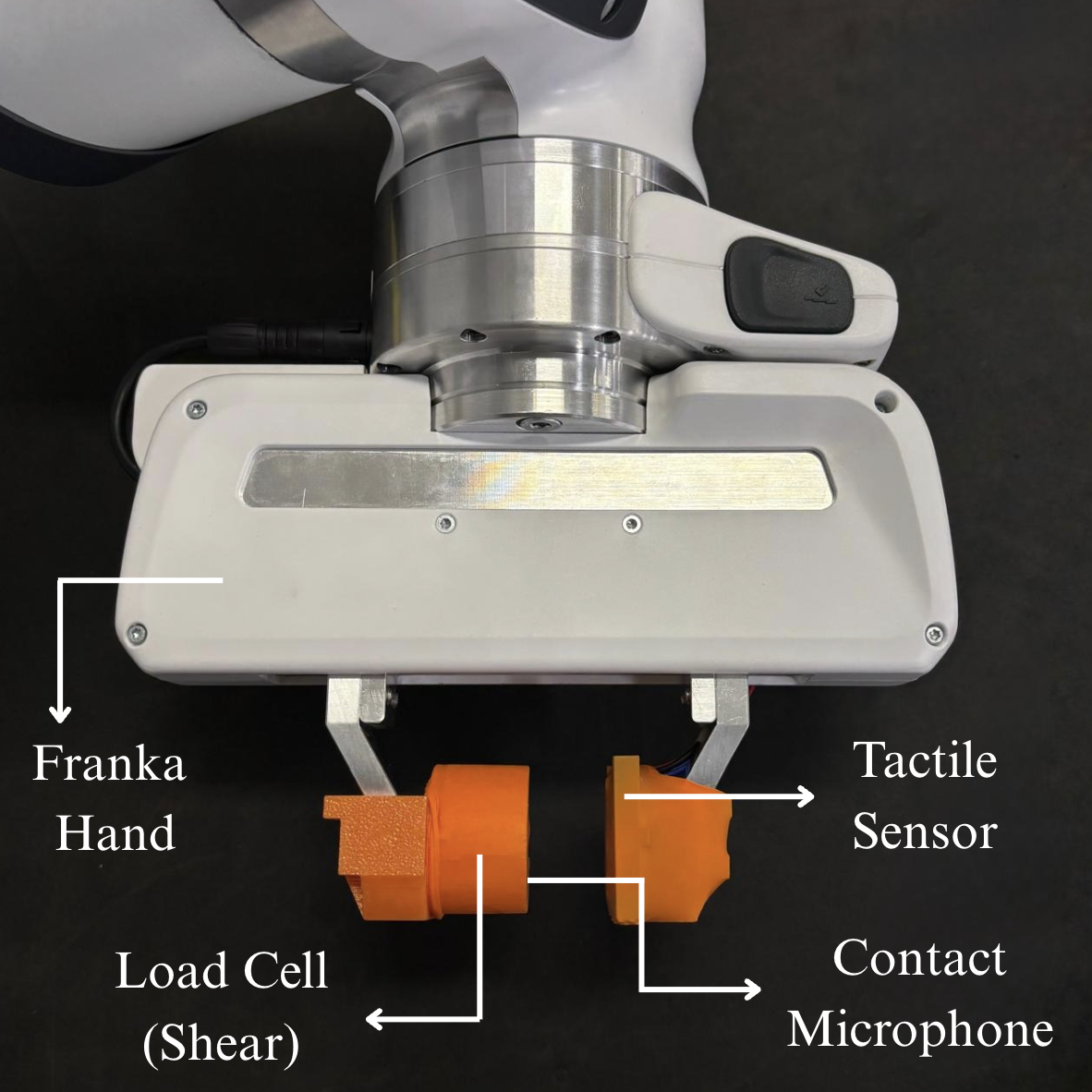}
    \caption{}
  \end{subfigure}
  \caption{Sensorized Franka Hand used for data collection. In (a), the gripper directly pushes the component into its mating counterpart while in (b), the gripper holds the part and performs the insertion into its mating counterpart.}
  \label{fig:franka_hand}
\end{figure}

We instrumented the setup with multiple sensors (Fig.~\ref{fig:franka_hand}), each sampled at $100$\,Hz, to capture snap events from complementary perspectives:
\begin{enumerate}[label=\alph*.]
    \item \emph{Robot Joint States:} Proprioceptive data were recorded via the native robot API, including joint positions, velocities, and torques across all seven degrees of freedom.
    \item \emph{Uniaxial Load Cell (Futek LLB\,300):} The load cell was mounted in two configurations—first between the Franka Hand and the part to capture lateral (shear) forces during snap trials, and later reoriented to align with the insertion axis for direct axial force measurement. 
    \item \emph{Contact Microphones:} Two high-sensitivity contact microphones (bandwidth: $20$–$20{,}000$\,Hz) were used: one attached to the gripper body and another at the interface between the gripper and the mating parts. The interface-mounted sensor consistently recorded high-frequency transients characteristic of snap engagements, while the gripper-mounted sensor showed weak or inconsistent responses.
    \item \emph{Accelerometer (ADXL\,355, $\pm8\,g$):} Mounted on the gripper body to record impulsive inertial responses. It captured sharp accelerations coinciding with snap transitions, serving as a vibration-based proxy for contact dynamics.
    \item \emph{Tactile Sensors:} Thin force-sensitive resistor (FSR) pads were embedded in the inner gripper fingers to monitor surface pressure distribution. Although not direct snap detectors, they exhibited abrupt pressure changes at engagement.
\end{enumerate}
This multimodal dataset provides a synchronized, diverse basis for evaluating the reliability of different sensing modalities in detecting snap-fit events. The complete dataset, covering the bold-faced object categories from Table~\ref{tab:taxonomy}, is made publicly available for research use.

\begin{figure*}[ht]
  \centering
  \begin{subfigure}{0.3\linewidth}
    \centering
    \includegraphics[width=\textwidth]{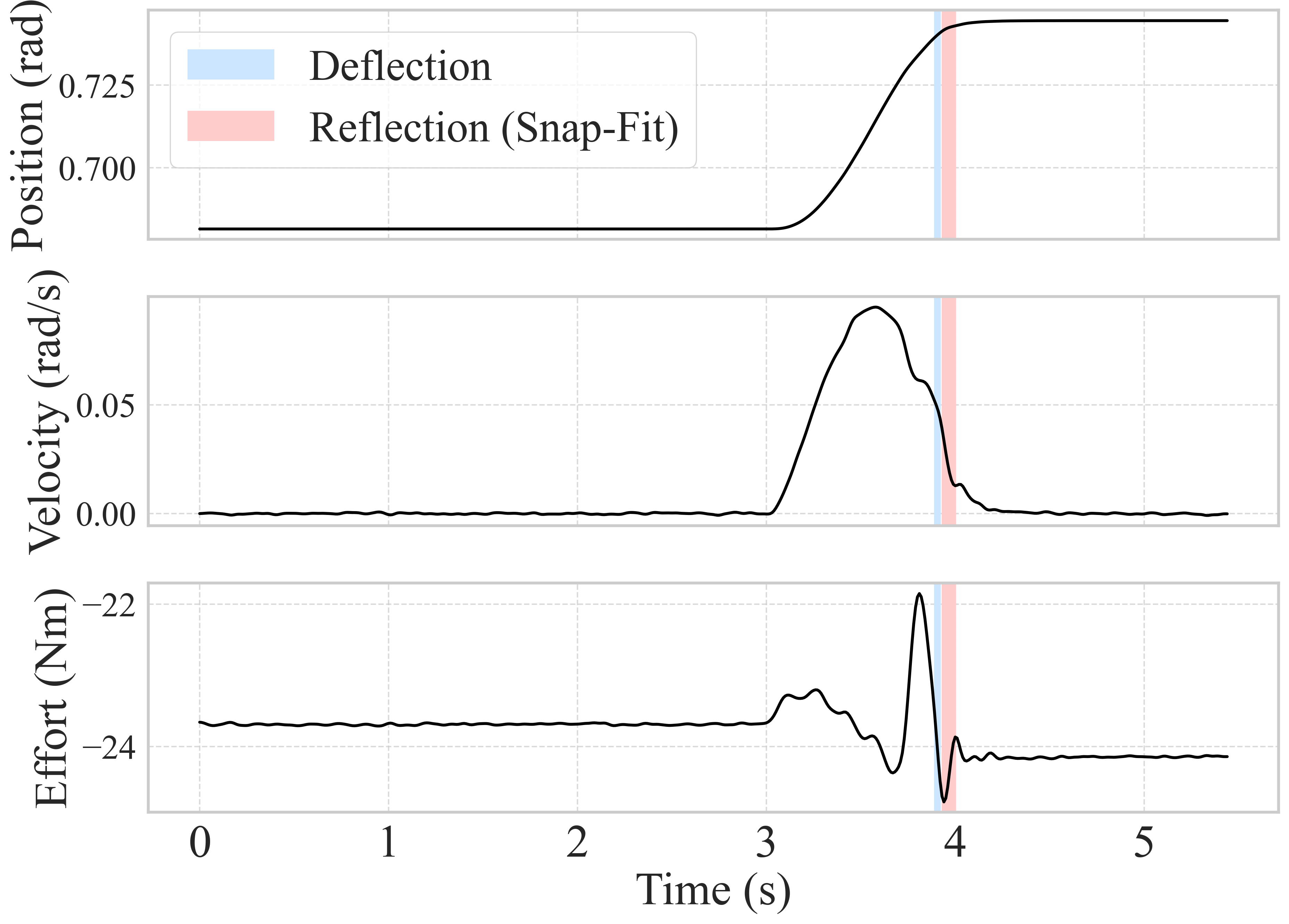}
    \caption{Bottle Cap}
  \end{subfigure}
  \hfill
  \begin{subfigure}{0.3\linewidth}
    \centering
    \includegraphics[width=\textwidth]{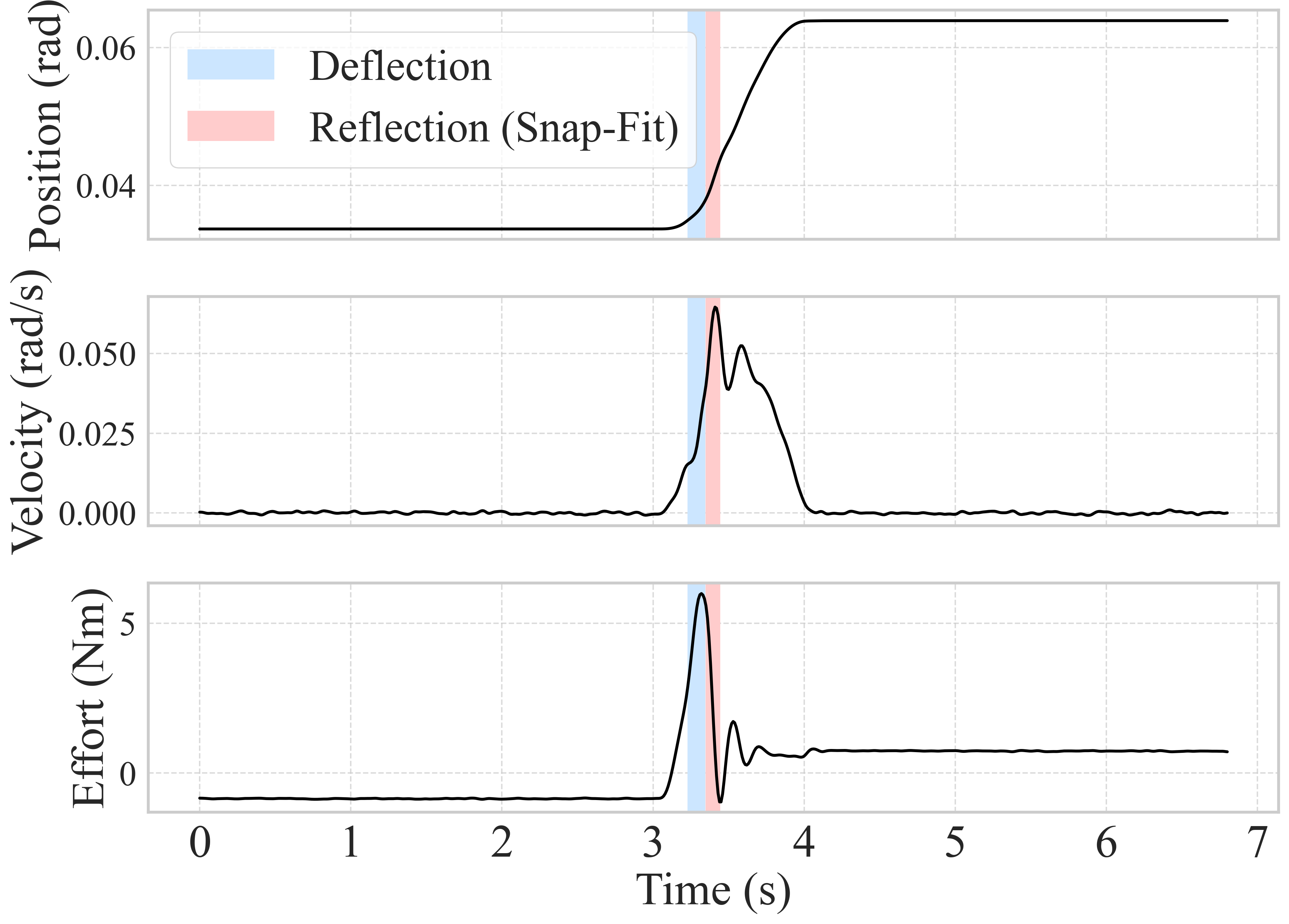}
    \caption{Type-C USB Cable}
  \end{subfigure}
  \hfill
  \begin{subfigure}{0.3\linewidth}
    \centering
    \includegraphics[width=\textwidth]{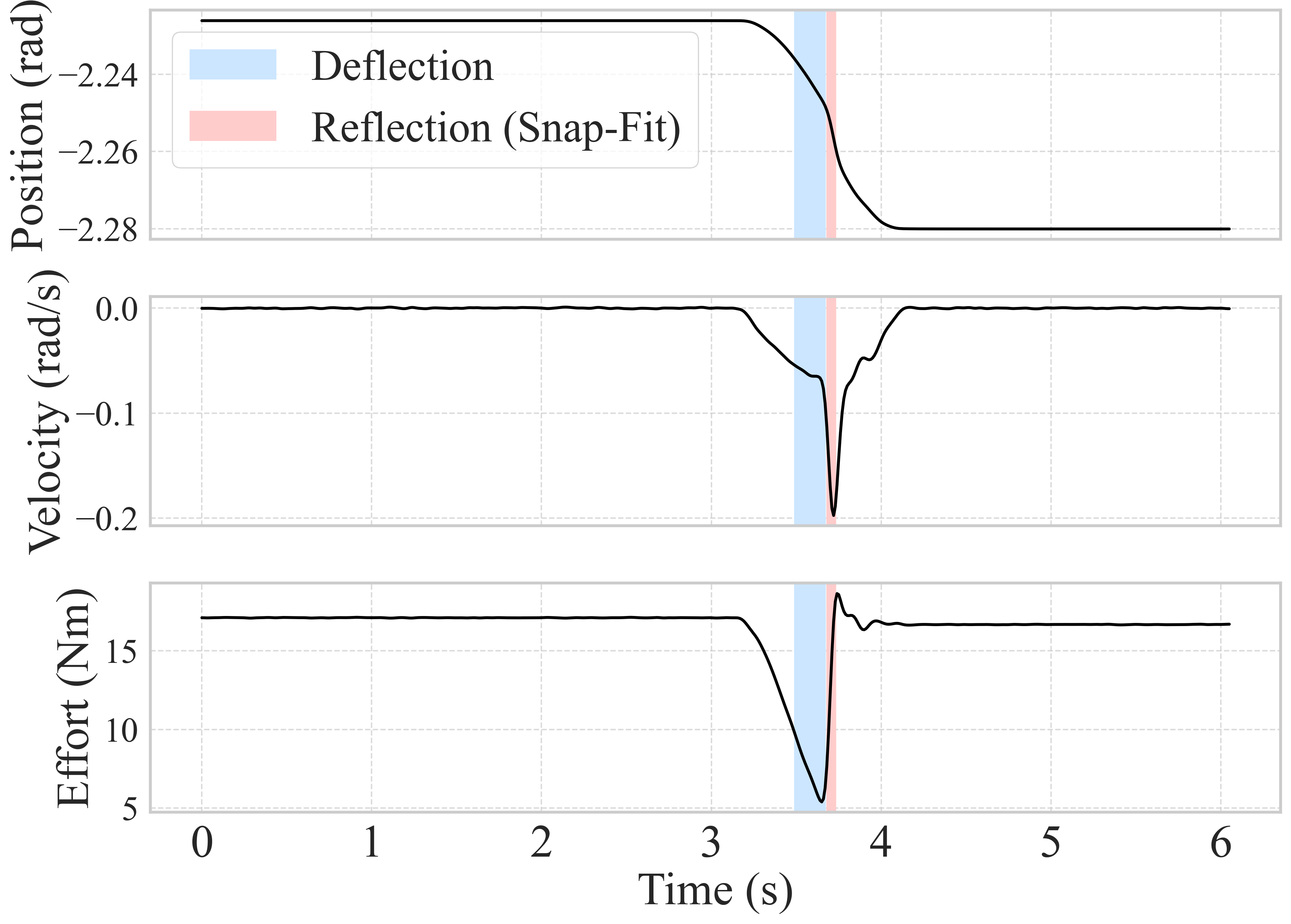}
    \caption{Highlighter Cap}
  \end{subfigure}

  \vspace{0.5cm}

  \begin{subfigure}{0.3\linewidth}
    \centering
    \includegraphics[width=\textwidth]{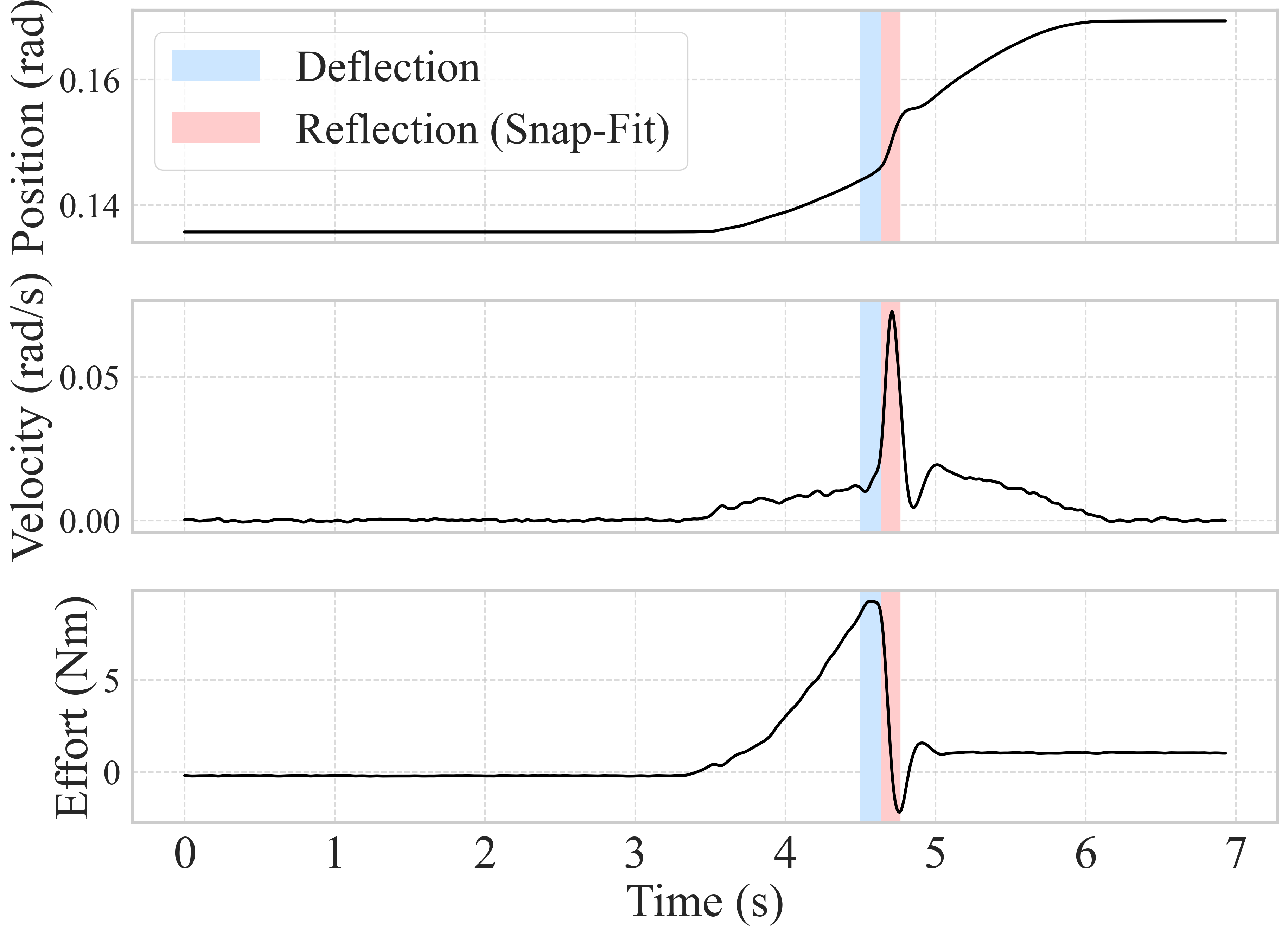}
    \caption{Lens and Frame}
  \end{subfigure}
  \hfill
  \begin{subfigure}{0.3\linewidth}
    \centering
    \includegraphics[width=\textwidth]{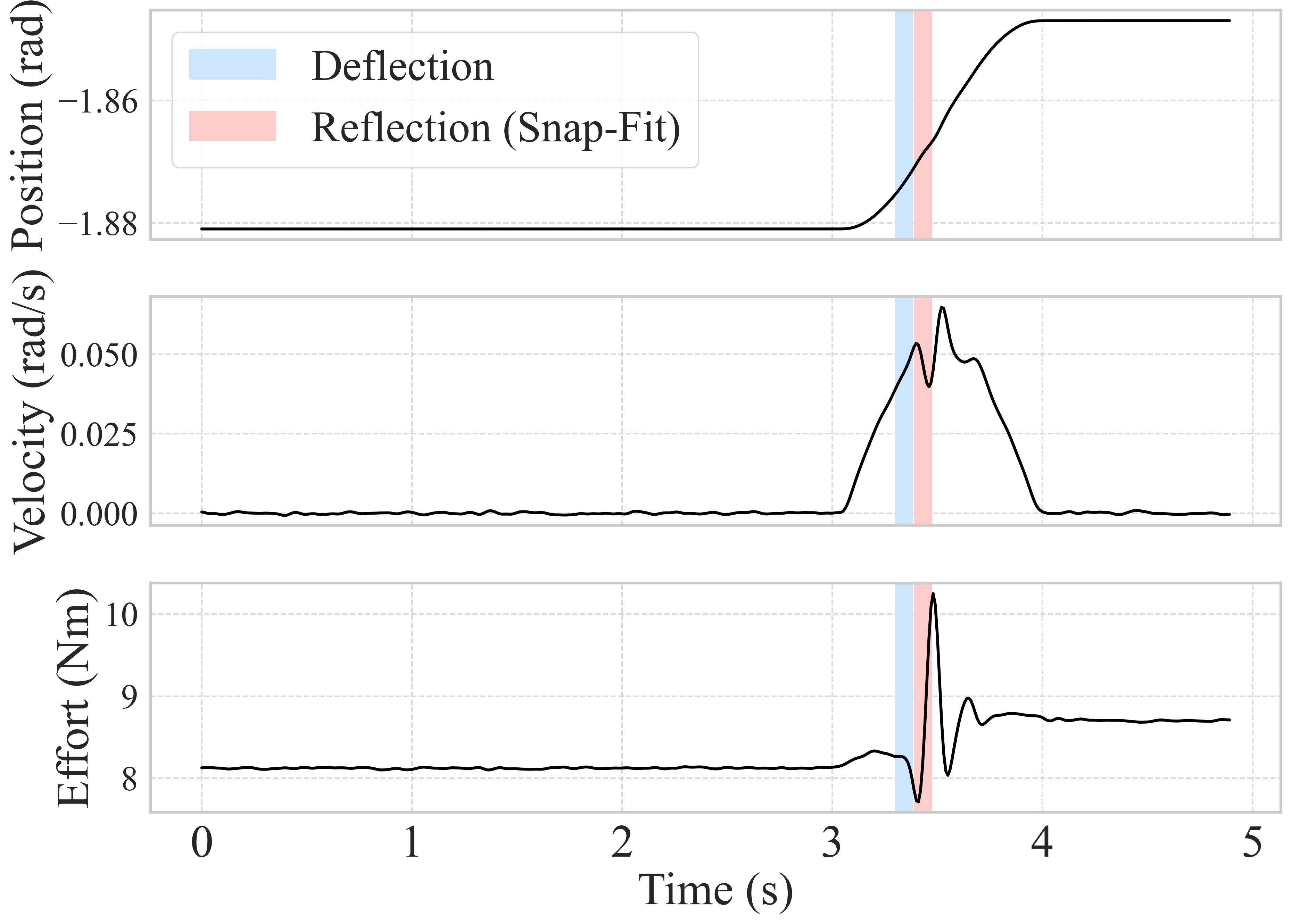}
    \caption{Custom Part}
  \end{subfigure}
  \hfill
  \begin{subfigure}{0.3\linewidth}
    \centering
    \includegraphics[width=\textwidth]{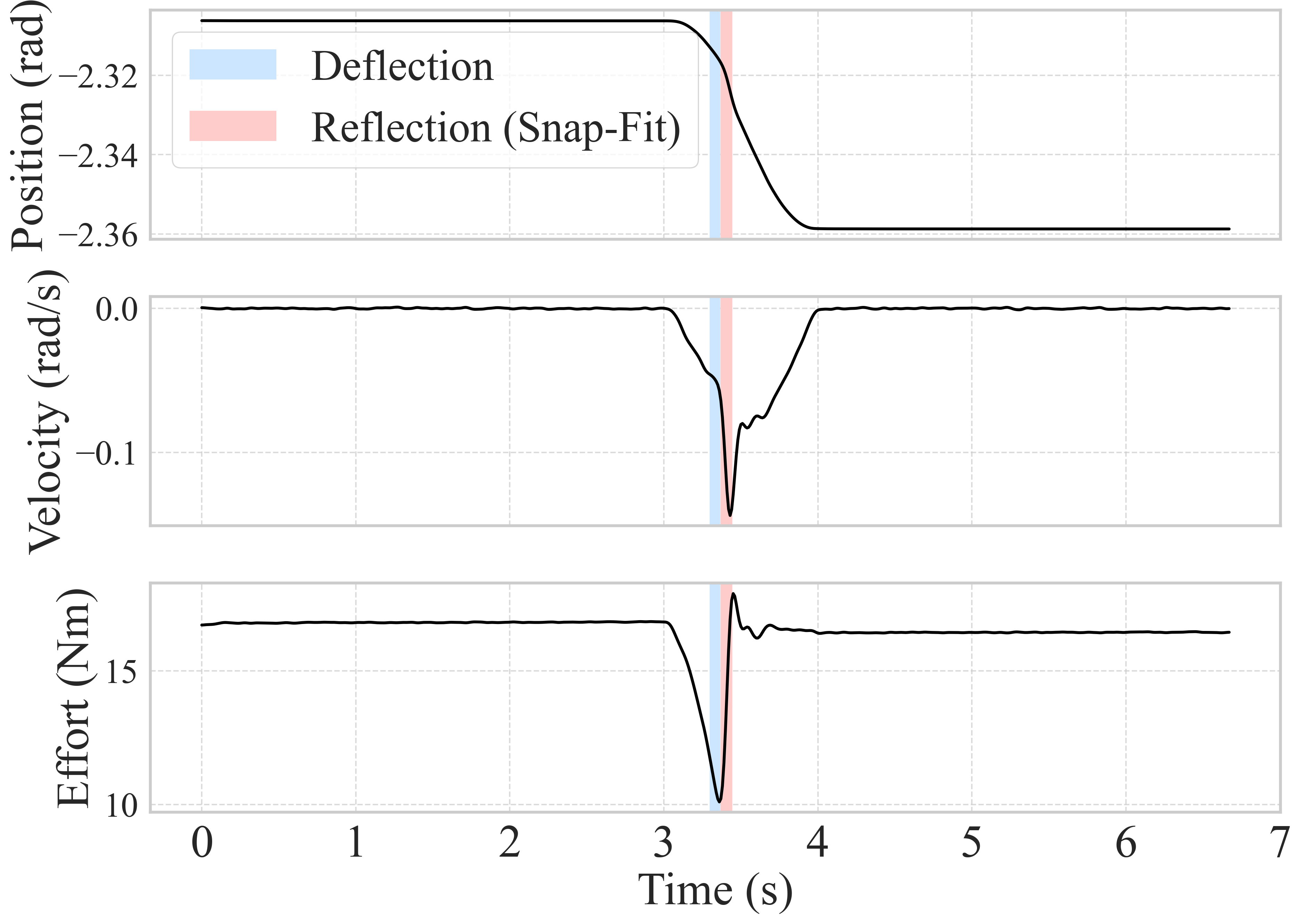}
    \caption{Marker Cap}
  \end{subfigure}

  \caption{Snap signatures noted on the most prominent joints of Franka FR3.}
  \label{fig:snap_objects}
\end{figure*}

\subsubsection{Sensor Benchmarking}
To enable uniform comparison across sensing modalities, we trained a common one-dimensional convolutional neural network (1D-CNN) on sliding windows of raw time-series signals. CNNs were selected for their ability to extract temporal features directly from unprocessed data, thereby avoiding modality-specific feature engineering and ensuring consistency across sensors. Table~\ref{tab:sensor_benchmark} reports accuracy, precision, recall, and $F_1$-score for each modality (metrics defined explicitly in \S~\ref{subsubsec:model}).

Joint velocity achieved the highest overall performance ($F_1=0.8822$), combining strong precision and recall, while also being universally available on robotic platforms. Joint effort and axial force also performed well ($F_1=0.8562$ and $0.8677$, respectively), reaffirming that both kinematic and force cues contain strong engagement signatures. In contrast, shear force, tactile, and acoustic signals yielded lower recall, reflecting limited sensitivity to snap events that vary in direction or contact location.

\begin{table}[ht]
\centering
\caption{Performance metrics across different sensor modalities using a 1D-CNN classifier.}
\label{tab:sensor_benchmark}
\begin{tabular}{lcccc}
\toprule
\textbf{Sensor Modality}    & \textbf{Accuracy} & \textbf{Precision} & \textbf{Recall} & \textbf{$F_1$ score} \\
\midrule
Joint Velocity      & 0.9211   & 0.9233    & 0.8447 & 0.8822 \\
Joint Effort        & 0.9169   & 0.8799    & 0.8339 & 0.8562 \\
Load Cell (Axial)   & 0.9297   & 0.8935    & 0.8435 & 0.8677 \\
Tactile Sensor      & 0.8121   & 0.7647    & 0.6324 & 0.6922 \\
Load Cell (Shear)   & 0.7943   & 0.7079    & 0.5447 & 0.6156 \\
Contact Microphone  & 0.8874   & 0.8461    & 0.6252 & 0.7190 \\
Accelerometer       & 0.8529   & 0.8143    & 0.7989 & 0.8065 \\
\bottomrule
\end{tabular}
\end{table}

While these results reflect strong offline performance under fixed-length windowing, the same CNN models performed inconsistently during streaming inference. Offline evaluation relied on pre-segmented data windows manually aligned around engagement events, ensuring that the snap transient was fully contained within each sample. In real operation, however, the timing of engagement is unknown, and rolling windows often include only partial or shifted transients, leading to false or delayed detections. This limitation is well-documented in online time-series classification, where convolutional models lack temporal context to reconcile overlapping segments~\cite{bao2004detecting, baghezza2021offline}. To address this, we present \textit{SnapNet}, a temporally adaptive model for real-time proprioceptive snap detection. Motivated by its strong performance and universal availability across robotic platforms, we base SnapNet on joint-velocity signals.

\subsubsection{Model Architecture}
\label{subsubsec:model}
Let the input be $\mathbf{V} \in \mathbb{R}^{T \times N}$, representing normalized joint velocities over a window of $T$ time steps for $N$ joints. The time series for joint $n$ is denoted as
\[
\mathbf{v}^{(n)} = [v^{(n)}(1), v^{(n)}(2), \dots, v^{(n)}(T)]^\top \in \mathbb{R}^{T}.
\]

\begin{enumerate}[label=\alph*.]
    \item \emph{Per-Joint CNN Encoder:} Each $\mathbf{v}^{(n)}$ is processed by a shared 1D convolutional encoder $f_{\text{CNN}}(\cdot)$ with kernel size $k$ to extract local temporal features:
    \[
        \mathbf{x}^{(n)} = f_{\text{CNN}}(\mathbf{v}^{(n)}) = [x_1^{(n)}, \dots, x_{T'}^{(n)}], \quad x_t^{(n)} \in \mathbb{R}^{d_c},
    \]
    where $T'$ is the reduced sequence length and $d_c$ is the feature dimension, both treated as hyperparameters. Convolutional filters followed by rectified linear unit (ReLU) activation capture joint-independent transients such as sharp velocity spikes associated with snap engagement.

    \item \emph{Per-Joint GRU Encoder:} Each feature sequence $\mathbf{x}^{(n)}$ is passed through a gated recurrent unit (GRU) encoder $f_{\text{GRU}}(\cdot)$  to model temporal dependencies. The GRU produces a hidden state at each time step $t \in [1,\, T']$, the final hidden state ${h}^{(n)}_{T'}$ is used as the joint-level embedding, denoted simply as ${h}^{(n)}$.
        \[
        h^{(n)} = f_{\text{GRU}}(\mathbf{x}^{(n)})_{T'} \in \mathbb{R}^{d_h},
    \]
    where $d_h$ is the GRU hidden dimension and the final hidden state serves as the joint-level embedding. The CNN–GRU stack ensures that brief but discriminative local patterns contribute to the temporally aggregated representation.

    \item \emph{Attention Pooling Across Joints:} 
    A global embedding is obtained via attention over joint embeddings 
    $\{h^{(1)}, \dots, h^{(N)}\}$. 
    For each joint, an attention score $e^{(n)}$ and its weight 
    $\alpha^{(n)}$ are computed as:
    \[
    e^{(n)} = \mathbf{u}_a^\top \tanh(\mathbf{W}_a h^{(n)} + \mathbf{b}_a), \quad
    \alpha^{(n)} = \text{softmax}(e^{(n)}).
    \]
    where $\mathbf{W}_a, \, \mathbf{b}_a$, and $\mathbf{u}_a$ are trainable attention parameters.
    The weighted sum
    $\mathbf{h}_{\text{global}} = \sum_{n=1}^{N} \alpha^{(n)} h^{(n)}$
    yields a permutation-invariant representation that emphasizes joints exhibiting distinct snap-like transients.
    
    \item \emph{Classification Layer:} 
    The global embedding is passed through a linear layer with sigmoid activation to estimate snap probability:
    \[
    p = \sigma(\mathbf{w}_o^\top \mathbf{h}_{\text{global}} + b_o),
    \]
    where $\mathbf{w}_o \in \mathbb{R}^{d_h}$ and $b_o \in \mathbb{R}$. 
\end{enumerate}

\begin{figure}[h]
    \centering
    \includegraphics[width=\linewidth]{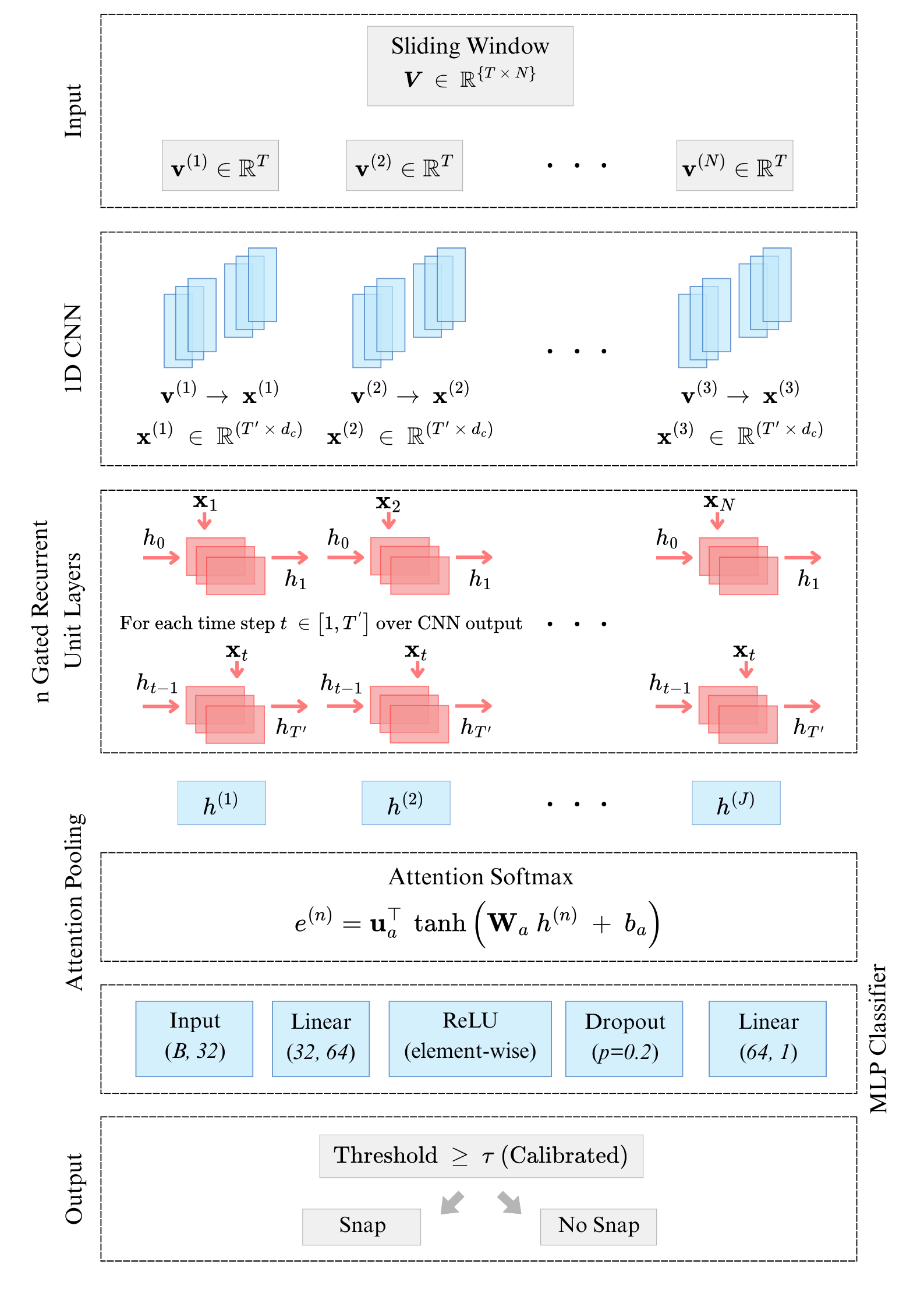}
    \caption{\textbf{SnapNet architecture.} Per-joint CNN-GRU encoders extract local and temporal features from joint velocities, which are fused via attention and classified through a sigmoid layer.}
    \label{fig:snapnet_arch}
\end{figure}

To address class imbalance and emphasize rare engagement events, we employ the Focal Loss~\cite{lin2017focal}, a variant of the binary cross-entropy loss that down-weights easy negatives and focuses training on hard examples. For a ground-truth label $y \in \{0,1\}$ and predicted probability $p$, the loss is
\[
\mathcal{L}_{\text{focal}}(y, p) = -\alpha\, y (1 - p)^{\gamma} \log(p)
                                  - (1 - \alpha)(1 - y) p^{\gamma} \log(1 - p),
\]
where $\alpha$ balances class weights and $\gamma$ controls the focusing strength.

A detection threshold $\tau$ is calibrated by sweeping its value on the validation set and selecting $\tau^\star$ that maximizes the $F_1$-score:
\[
F_1(\tau) = \frac{2 \cdot \text{Precision}(\tau) \cdot \text{Recall}(\tau)}
                 {\text{Precision}(\tau) + \text{Recall}(\tau)}.
\]
with $\text{Precision} = \frac{TP}{TP + FP}, \quad
\text{Recall} = \frac{TP}{TP + FN}$
where TP denotes true positives and TN denotes true negatives.
The fixed $\tau^\star$ is then applied during deployment to ensure stable and consistent snap detections.

\subsubsection{Results}
We trained SnapNet for $500$ epochs using focal loss ($\alpha=0.25$, $\gamma=2.0$) on sliding windows of length $T=50$ time steps ($500$\,ms at $100$\,Hz). Optimization was performed using the Adam optimizer (learning rate $1\times10^{-3}$) with a batch size of $64$. To mitigate class imbalance between snap and no-snap windows, we calibrated the decision threshold~$\tau$ on the validation set to maximize $F_1$, yielding an optimal value of $\tau^\star = 0.45$.

Table~\ref{tab:method_comparison} presents a side-by-side comparison of SnapNet with the SVM-based method~\cite{doltsinis2019machine} and the recent R-RNN model~\cite{cui2023fast}. SnapNet achieves an $F_1$ score of $0.9778$, substantially outperforming the SVM baseline ($F_1 = 0.7692$) and closely matching the performance of the more computationally intensive R-RNN model ($F_1 = 0.9729$). Although differences in dataset construction and windowing strategies make exact numerical comparisons challenging, these results suggest that a compact CNN-GRU-attention architecture can reliably capture snap-fit signatures using proprioceptive sensing alone.

\begin{table}[t]
  \centering
  \caption{Result comparison of snap-fit detection with the SVM and R-RNN based methods (offline performance evaluated using an $80$–$10$–$10$ train-validation-test split)}
  \label{tab:method_comparison}
  \resizebox{\columnwidth}{!}{%
    \begin{tabular}{l c c c c c c}
      \toprule
      Item                        & TP   & TN      & Accuracy & Recall  & Precision & $F_1$     \\
      \midrule
      SVM~\cite{doltsinis2019machine}     &   25 &    365   & 0.9630 & 1.0000 & 0.6250 & 0.7692 \\
      R-RNN~\cite{cui2023fast}            &  987 &  16\,655 & 0.9969 & 0.9811 & 0.9648 & 0.9729 \\
      \textbf{SnapNet}                    &  440 &   6\,590 & 0.9972 & 0.9778 & 0.9778 & 0.9778 \\
      \bottomrule
    \end{tabular}%
  }
\end{table}

\begin{figure}[t]
    \centering
    \includegraphics[width=0.95\linewidth]{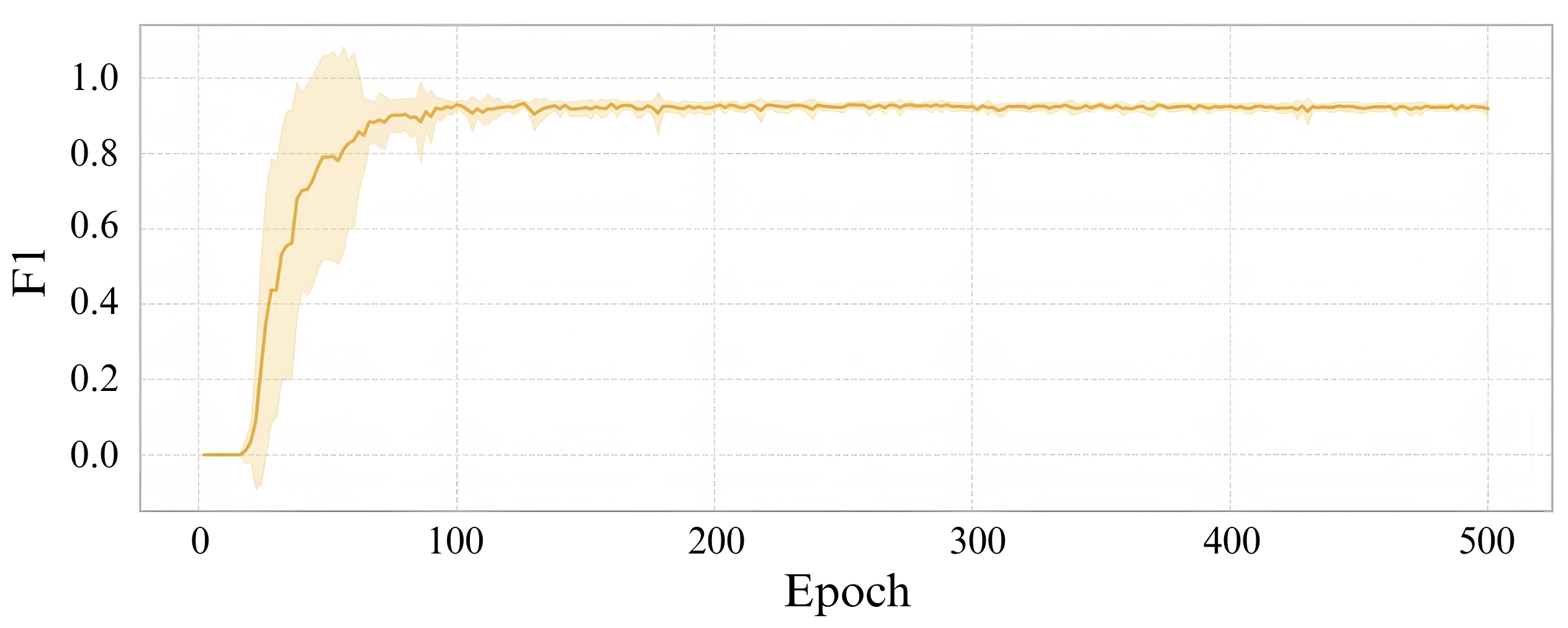}
    \caption{Test $F_1$ score versus training epoch for SnapNet.}
    \label{fig:learning_curves}
\end{figure}

Figure~\ref{fig:learning_curves} illustrates the evolution of SnapNet’s test $F_1$ over $500$ epochs (mean $\pm$ std across five folds), showing rapid convergence by epoch $30$ and stable performance thereafter. 

To understand the contribution of individual components within SnapNet, we performed a systematic ablation by selectively disabling the CNN encoder, GRU encoder, and attention module. These correspond to local transient extraction, temporal sequence modeling, and joint-level relevance weighting, respectively. Table~\ref{tab:ablation} reports accuracy, precision, recall, and $F_1$ for each variant.

Removing the attention module results in a notable performance drop ($F_1 = 0.9017$), indicating its importance in emphasizing joints that express clearer snap signatures. Removing either the CNN or GRU leads to further degradation ($F_1 = 0.8568$ and $0.8736$), showing that both short- and long-horizon temporal features are essential for distinguishing snap events from routine arm motion. Together, these results confirm that SnapNet’s full architecture is synergistic, with all three components contributing meaningfully to its detection performance. The experimental results from the real-time deployment of SnapNet are presented in (\S\ref{subsec:res_detection}).

\begin{table}[t]
\centering
\caption{Ablation on SnapNet components.}
\label{tab:ablation}
\vspace{2mm}
\begin{tabular}{lcccc}
\toprule
\textbf{Model Variant} & \textbf{Accuracy} & \textbf{Precision} & \textbf{Recall} & \textbf{F1} \\
\midrule
SnapNet  & 0.9972 & 0.9778 & 0.9778 & 0.9778 \\
w/o Attention        & 0.9874 & 0.9122 & 0.8915 & 0.9017 \\
w/o GRU              & 0.9842 & 0.8841 & 0.8633 & 0.8736 \\
w/o CNN              & 0.9823 & 0.8714 & 0.8427 & 0.8568 \\
\bottomrule
\end{tabular}
\end{table}

\subsection{Bimanual Coordination via Dynamical Systems}
\label{subsec:ds}

We formulate bimanual coordination within a dynamical systems (DS) framework, where each arm follows a phase-indexed reference trajectory that evolves smoothly over time. The formulation extends the phase-coupled DS model of~\cite{KhadivarAdaptiveFingers} by incorporating adaptive gain scheduling and task-dependent coupling, enabling seamless switching between synchronized transport and decoupled insertion phases.

\subsubsection{Phase and Path Parameterization}
Each arm $i$ progresses along its motion path according to a normalized phase variable $z_i(t) \in [0, 1]$, defined as:
\begin{equation}
  z_i(t) = \frac{\|\mathbf{p}_i - \mathbf{p}_i^{(1)}\|}{\|\mathbf{p}_i^{(2)} - \mathbf{p}_i^{(1)}\|},
\end{equation}
where $\mathbf{p}_i^{(1)}$ and $\mathbf{p}_i^{(2)}$ are the start and end points of the desired trajectory segment, and $\mathbf{p}_i(t)$ is the current end-effector position.  
The phase-indexed reference trajectory is given by:
\begin{equation}
  \boldsymbol{\gamma}_i(z_i) = \mathbf{p}_i^{(1)} + z_i \bigl( \mathbf{p}_i^{(2)} - \mathbf{p}_i^{(1)} \bigr),
\end{equation}
which generalizes naturally to nonlinear paths through appropriate parameterization.

\subsubsection{Coupled Phase Dynamics}
Coordination between the arms is achieved by coupling their phase dynamics.  
Let the average phase be:
\[
z^c = \frac{1}{2}(z_1 + z_2).
\]
Further, let the evolution of each arm’s phase be governed by:
\begin{equation}
  \dot{z}_i = g_i(z_i) + \kappa\, g_c(z_i),
  \label{eq:z_dyn_main}
\end{equation}
where $\kappa \in \{0,1\}$ is a task-phase-dependent coupling flag that enables ($\kappa=1$) or disables ($\kappa=0$) synchronization within the same framework.  
The terms $g_i(z_i)$ and $g_c(z_i)$ define the self and coupling dynamics:
\begin{align}
  g_i(z_i) &= -k_1 \frac{z_i - z_i^{d}}{(|z_i - z_i^{d}| + \varepsilon)^{1 - \eta}}, \label{eq:intermediate1} \\
  g_c(z_i) &= -k_2 \frac{z_i - z^c}{(|z_i - z^c| + \varepsilon)^{1 - \eta}},
\label{eq:intermediate2}
\end{align}
where $z_i^{d}$ is the desired phase setpoint, and parameters $k_1, k_2 > 0$, $\varepsilon > 0$, and $\eta \in (0,1)$ govern convergence rate and smoothness.  
Thus, during transport ($\kappa=1$), both arms remain phase-locked and upon insertion ($\kappa=0$), they evolve independently.

\vspace{4pt}
\noindent\textit{Theorem 1:} 
Consider the dynamics defined by~\eqref{eq:z_dyn_main}, \eqref{eq:intermediate1}, and \eqref{eq:intermediate2}. Further, let $k_1,k_2>0$, $\varepsilon>0$, and $\eta\in(0,1)$. Then, the following statements hold.
\begin{enumerate}
\item[(a)] If $\kappa=0$ and the desired phase $z_i^d$ is constant, the equilibrium $z_i\equiv z_i^d$ is globally asymptotically stable,
\item[(b)] If $\kappa=1$ and $z_i^d\equiv z^d$ for all arms, the phases achieve global consensus,
\[
\lim_{t\to\infty} z_i(t)=z^{d}, \qquad
\lim_{t\to\infty}\bigl(z_i(t)-z^c(t)\bigr)=0.
\]
\end{enumerate}
\textit{Proof:} See Appendix~A \hfill $\blacksquare$ \\
\textit{Remark 1:} Compared to the intermediate phase dynamics~\cite{KhadivarAdaptiveFingers}, where a single bounded mapping governs evolution, our decomposition into $g_i$ and $g_c$
makes the coupling strength explicit and separable, allowing smooth modulation between independent and synchronized behavior.

\subsubsection{Task-Space Tracking with Adaptive Gain Scheduling}
Define the task-space tracking error $\mathbf{e}_i = \mathbf{p}_i - \boldsymbol{\gamma}_i(z_i)$.  
The velocity command for each arm is:
\begin{equation}
  \dot{\mathbf{p}}_i = -\alpha_i(E_i)\, \mathbf{e}_i, \quad \text{where } E_i = \|\mathbf{e}_i\|.
  \label{eq:x_dyn_main}
\end{equation}
The adaptive gain $\alpha_i(E)$ modulates responsiveness according to tracking error:
\begin{equation}
  \alpha_i(E) = \alpha_{\min} + (\alpha_{\max} - \alpha_{\min}) \frac{E}{E + \delta},
  \label{eq:adaptive_gain}
\end{equation}
with $0 < \alpha_{\min} < \alpha_{\max}$ and $\delta > 0$.  
This ensures strong corrective action for large deviations and smooth convergence near the target.  

\vspace{4pt}
\noindent\textit{Theorem 2:} Consider the tracking law \eqref{eq:x_dyn_main} with adaptive gain \eqref{eq:adaptive_gain}
with $0<\alpha_{\min}<\alpha_{\max}$ and $\delta>0$. Further, let $\boldsymbol{\gamma}_i(\cdot)$ be Lipschitz, and let $\lim_{t\to\infty} z_i(t) = z_i^d$ and $\lim_{t\to\infty} \dot z_i(t) = 0$. Then, 
\[
\lim_{t\to\infty}\mathbf{e}_i(t) = 0.
\]
\textit{Proof:} See Appendix~B \hfill $\blacksquare$

\subsection{Event-Triggerd Impedance Modulation}
\label{subsec:vic}

The mechanical analysis, as stated in \S\ref{subsubsec:mechanics}, explains that snap-fit engagement generates significant transient forces during the rapid energy release phase. A fixed high-stiffness control strategy would transmit these impact loads directly to the manipulators, potentially causing misalignment, part damage, or unpredictable contact behavior. Conversely, a permanently compliant strategy would lack the precision required for part alignment prior to engagement.

We therefore propose an event-triggered variable impedance control (VIC) strategy that preserves stiffness for precise alignment during approach, then rapidly reduces impedance upon snap detection to absorb impact loads. This approach integrates seamlessly with our bimanual pipeline: SnapNet's real-time detection of the snap event triggers a smooth, axis-selective modulation of Cartesian stiffness and damping.

We base our VIC implementation on the standard Cartesian impedance law \cite{hogan1985impedance}:
\begin{equation}
  \mathbf{F}_{\rm cmd}(t)
  = \mathbf{K}(t)\bigl(\mathbf{x}_{\rm ref}(t)-\mathbf{x}(t)\bigr)
  + \mathbf{D}(t)\bigl(\dot{\mathbf{x}}_{\rm ref}(t)-\dot{\mathbf{x}}(t)\bigr),
  \label{eq:vic_impedance}
\end{equation}
where \(\mathbf{x},\dot{\mathbf{x}}\) are the end-effector pose and velocity, \(\mathbf{x}_{\rm ref},\dot{\mathbf{x}}_{\rm ref}\) the DS-planner references, and \(\mathbf{K}(t),\mathbf{D}(t)\in\mathbb{R}^{3\times3}\) diagonal, positive-definite gain matrices.

Let $t_s$ denote the time of first snap detection, we then schedule the stiffness modulation as:
\begin{align}
  \mathbf{K}(t) &=
  \begin{cases}
    K_{0}, & t < t_s, \\
    K_{0} + \bigl(K_{\mathrm{f}} - K_{0}\bigr)\,e^{-\lambda (t - t_s)}, & t \ge t_s,
  \end{cases}
  \label{eq:vic_stiffness}\\
  \mathbf{D}(t) &= \alpha\,\sqrt{\mathbf{K}(t)},
  \label{eq:vic_damping}
\end{align}
where $\mathbf{K}_0$ and $\mathbf{K}_f$ are the stiffness levels before and after snap-fit engagement, $\lambda>0$ is the decay rate, and $\alpha>0$ scales damping relative to stiffness. 
The exponential stiffness transition~\eqref{eq:vic_stiffness} provides a smooth, first-order decay from $\mathbf{K}_0$ to $\mathbf{K}_f$, avoiding force discontinuities, while the proportional damping law~\eqref{eq:vic_damping} follows the critically-damped design convention that preserves passivity and prevents overshoot during rapid stiffness reduction.

Because $\mathbf{K}(t)$ and $\mathbf{D}(t)$ remain symmetric positive-definite and vary smoothly with bounded derivatives, the impedance law~\eqref{eq:vic_impedance} conforms to standard variable impedance control formulations known to preserve output passivity under continuous stiffness modulation~\cite{passive-vic}. 
This approach enables both precise pre-engagement alignment and compliant energy absorption during the critical snap transition phase.

\section{EXPERIMENTS}
\label{sec:experiments}

This section details the experimental setup and presents results for (i) snap-fit engagement detection on hardware, (ii) bimanual coordination via the proposed DS policy, and (iii) framework comparison of position control, constant-gain impedance, and event-triggered VIC.

\begin{figure*}[b]
  \centering
  \includegraphics[width=0.7\linewidth]{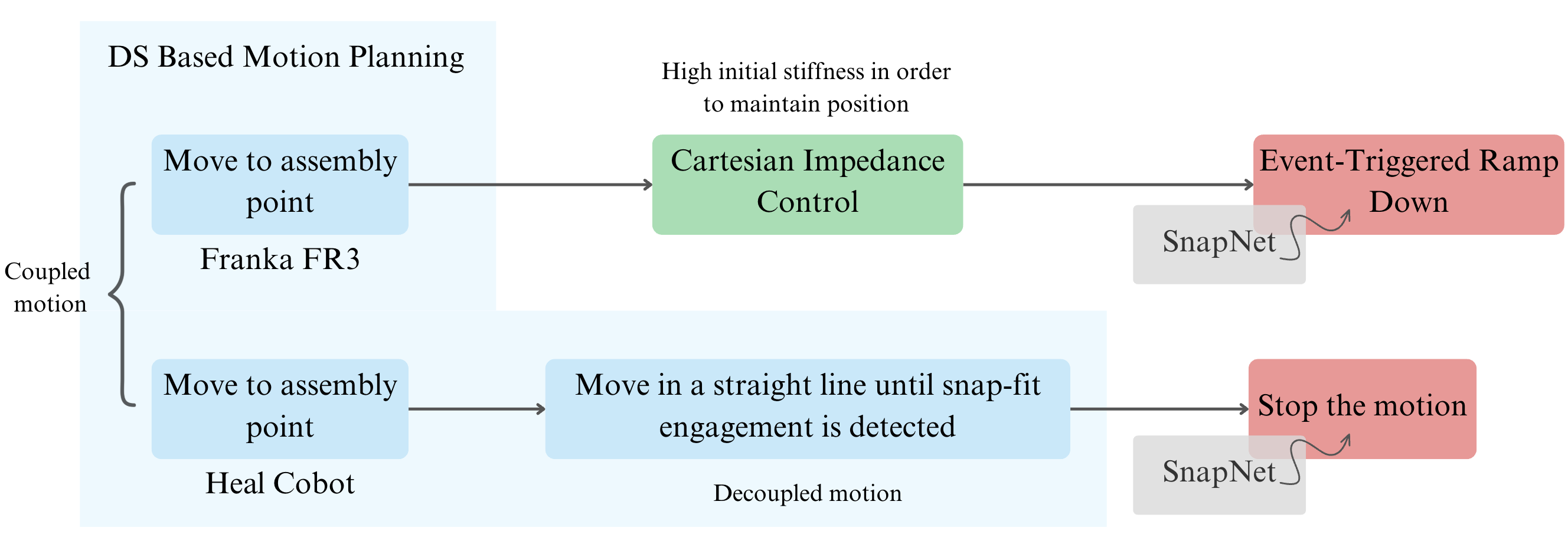}
  \caption{Proposed framework combining DS-based coordination, SnapNet detection, and event-triggered VIC. Coupled transport is followed by decoupled insertion, with impedance adaptation triggered by detected snap engagement.}
  \label{fig:framework_pipeline}
\end{figure*}

\subsection{Experimental Setup}
\label{subsec:setup}

Experiments were conducted on a heterogeneous bimanual platform consisting of a $7$-DoF Franka Emika FR3 and a $6$-DoF Addverb Heal cobot. We evaluated six snap-fit exemplars including a highlighter cap, marker-pen cap, water bottle-lid, Type-C USB Cable, lens-frame assembly, and an emergency-stop button. SnapNet was trained and deployed entirely on the Franka FR\,$3$ platform, using joint-velocity windows as input as detailed in Section~\ref{subsec:snapnet}.

The DS-based coordinator (\S~\ref{subsec:ds}) was executed in three phases. Let \( \mathbf{p}_i^{(j)} \in \mathbb{R}^d \) denote a sequence of keyframe positions for each arm, where \( j \in \{1, 2, 3, 4\} \). Specifically, \( \mathbf{p}_i^{(1)} \) corresponds to the initial rest pose before pick-up, \( \mathbf{p}_i^{(2)} \) represents the position at which the object is grasped, \( \mathbf{p}_i^{(3)} \) indicates the pre-assembly staging point, and \( \mathbf{p}_i^{(4)} \) denotes the final hold position for arm~$1$ and the approach point for arm~$2$.

The control is organized into phases indexed by \( \theta \in \{1,2,3\} \):
\begin{enumerate}[label=\alph*.]
  \item Phase $1$: Both arms move in a coupled manner from \( \mathbf{p}_i^{(1)} \to \mathbf{p}_i^{(2)} \) (pick-up).
  \item Phase $2$: The arms remain coupled and move to \( \mathbf{p}_i^{(3)} \) (transport and pre-alignment).
  \item Phase $3$: Arm~$1$ holds position at \( \mathbf{p}_1^{(4)} = \mathbf{p}_1^{(3)} \), while arm~$2$ completes the insertion from \( \mathbf{p}_2^{(3)} \to \mathbf{p}_2^{(4)} \). Coupling is disabled (\( \kappa = 0 \)) to allow unilateral motion.
\end{enumerate}
Phase transitions were triggered when all active arms satisfied $z_i>0.99$.

Upon SnapNet’s first detection time $t_s$, Cartesian stiffness along the insertion axis was exponentially reduced, with damping scheduled via criticality scaling (cf.~\eqref{eq:vic_stiffness} and \eqref{eq:vic_damping}). We used $K_{0} = 2000\,\text{N/m}$, $K_f = 100\,\text{N/m}$, $\lambda = 10\,\text{s}^{-1}$, and $\alpha = 2.0$. Concurrently, $\kappa \to 0$ to decouple the arms for compliant insertion.

The above components form an integrated framework for bimanual snap-fit assembly. DS-based motion planning generates coordinated transport trajectories, SnapNet provides real-time event detection from joint-velocity signatures, and event-triggered VIC modulates impedance to absorb impact loads during engagement. Figure~\ref{fig:framework_pipeline} illustrates the execution pipeline: both arms first move in coupled motion to the assembly point, the FR3 maintains position under high stiffness, and the Heal cobot performs the insertion. Upon detection of snap engagement, SnapNet triggers (i) a ramp-down of Cartesian stiffness on the holding arm and (ii) motion termination on the inserting arm, ensuring both alignment precision and compliant energy dissipation.

\begin{figure}[htbp]
  \centering
  \begin{subfigure}{0.45\linewidth}
    \centering
    \includegraphics[width=\textwidth]{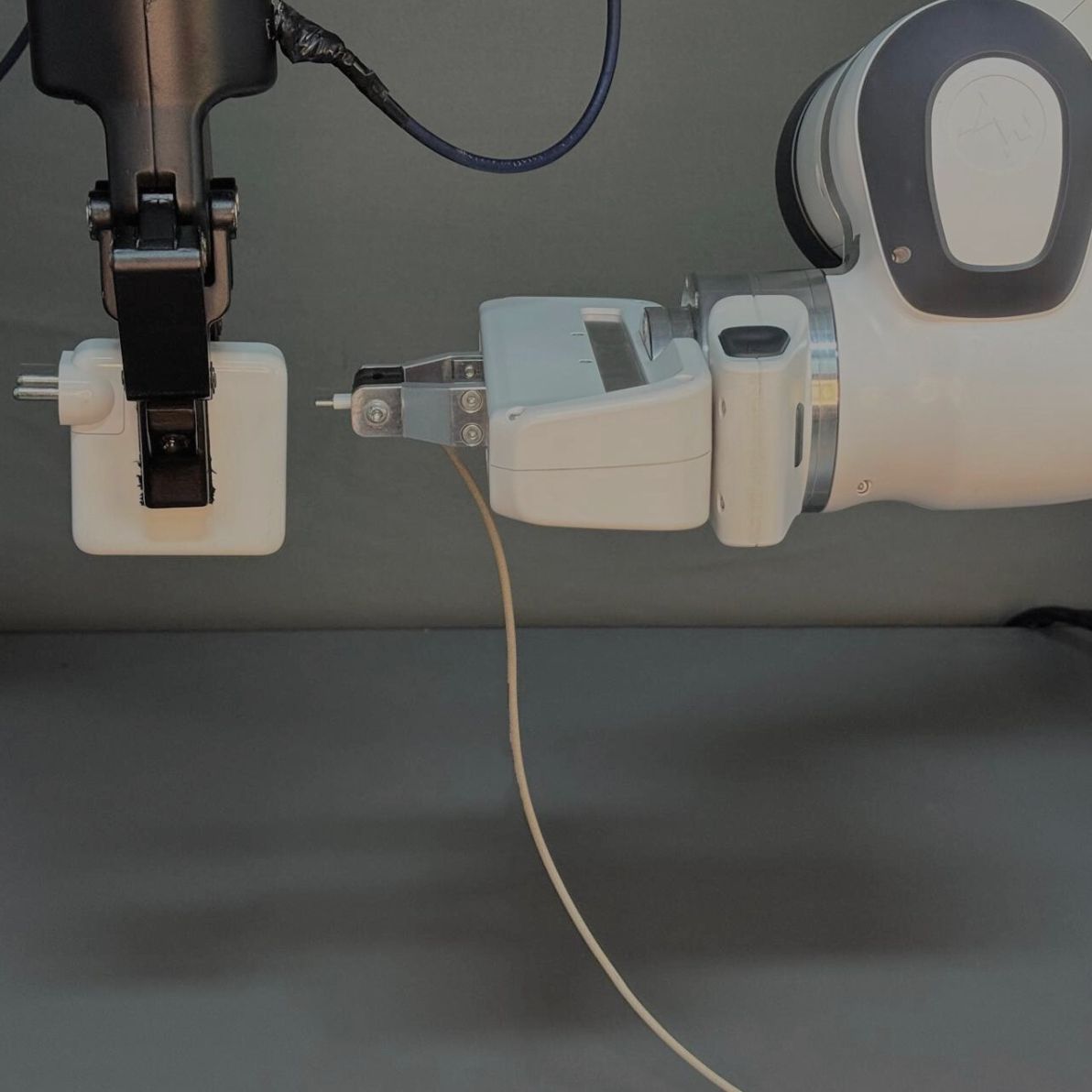}
  \end{subfigure}
  \hspace{0.5em}
  \begin{subfigure}{0.45\linewidth}
    \centering
    \includegraphics[width=\textwidth]{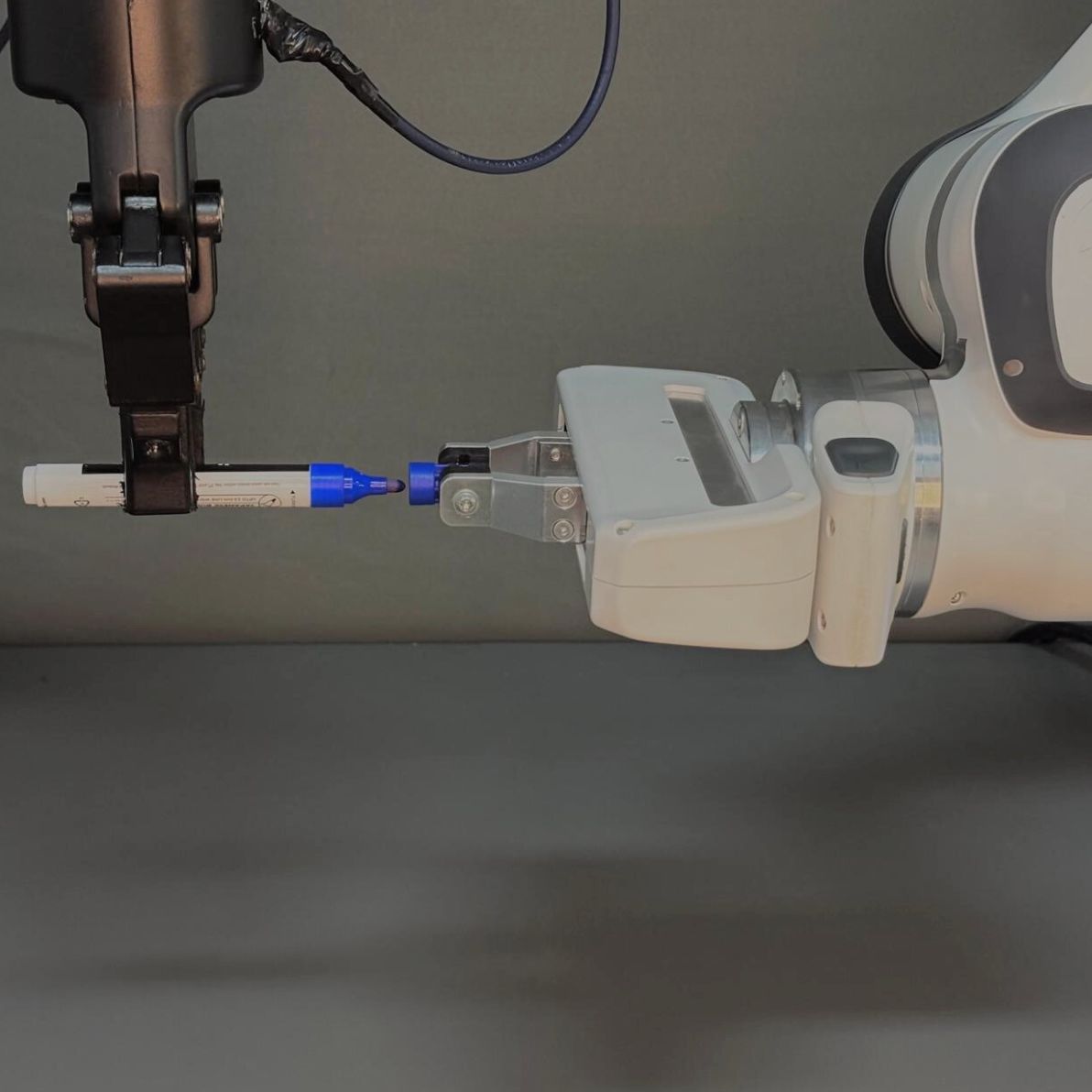}
  \end{subfigure}

    \vspace{1em}

  \begin{subfigure}{0.45\linewidth}
    \centering
    \includegraphics[width=\textwidth]{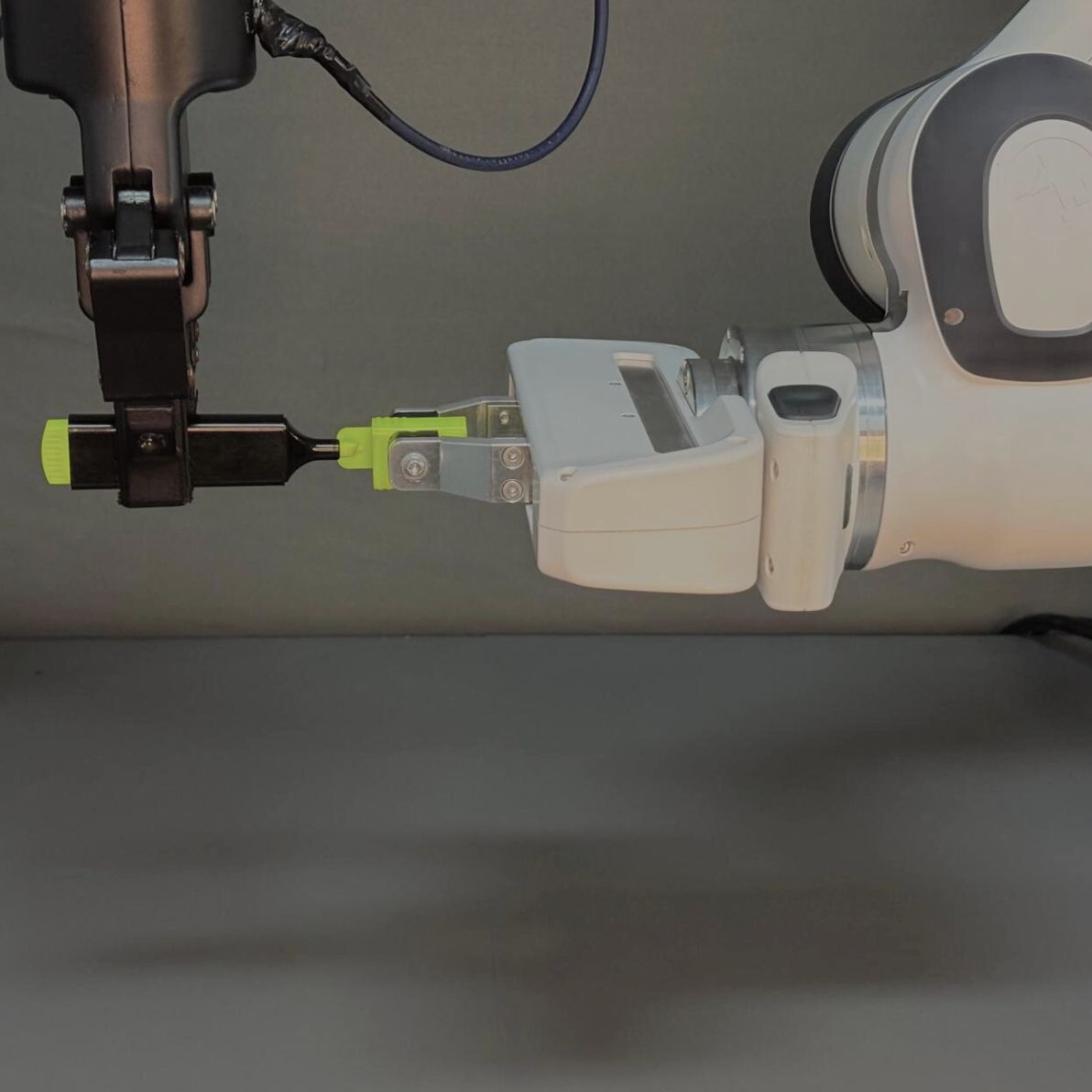}
  \end{subfigure}
  \hspace{0.5em}
  \begin{subfigure}{0.45\linewidth}
    \centering
    \includegraphics[width=\textwidth]{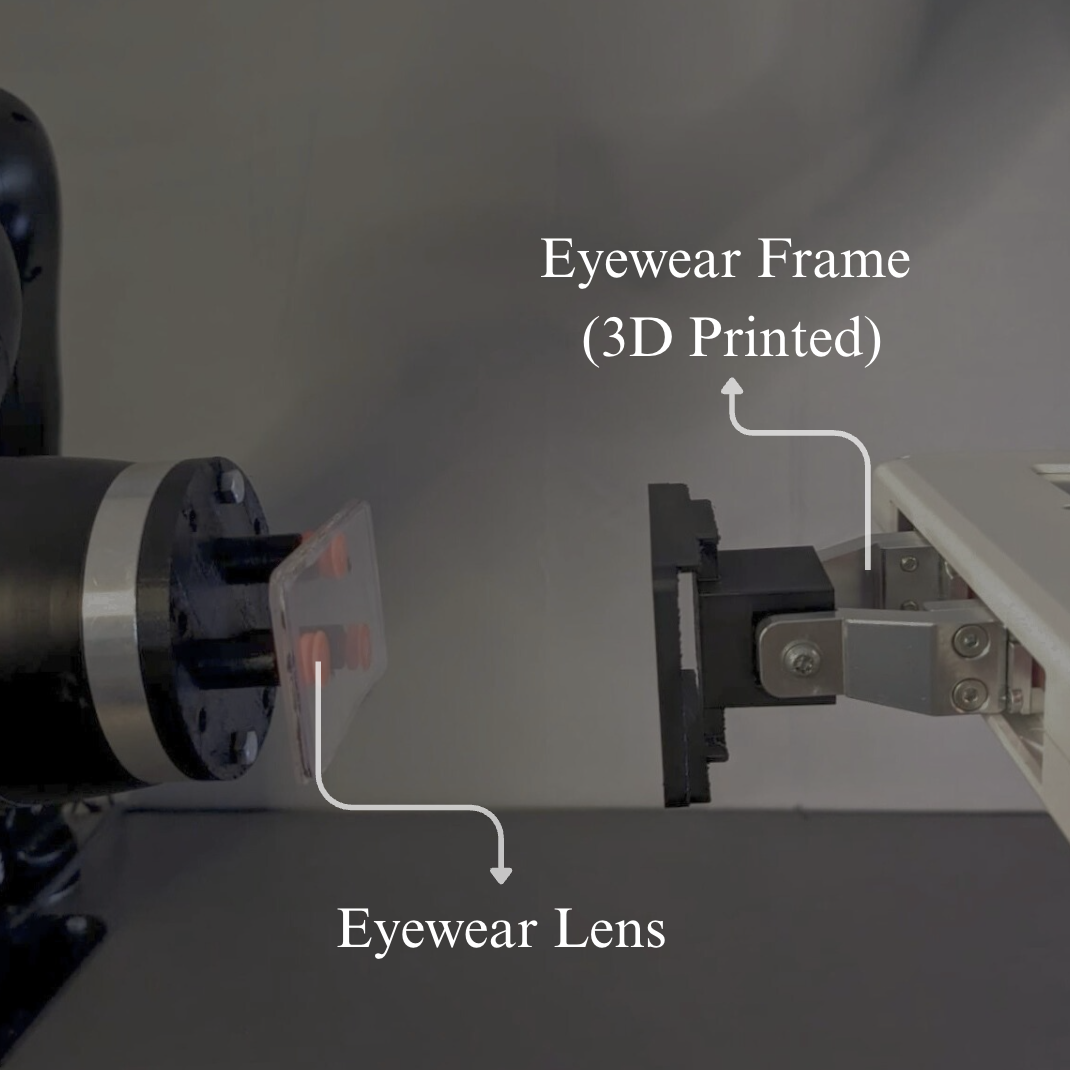}
  \end{subfigure}
  \caption{Representative bimanual snap-fit tasks: marker cap, lens-frame, highlighter cap, 
  Type-C USB Cable.}
  \label{fig:dual_snapfits}
\end{figure}

\subsection{Results}
\label{subsec:results}

\subsubsection{Snap-Fit Detection on Hardware}
\label{subsec:res_detection}
We report the fraction of true snap events detected by SnapNet over $15$ trials per part. Detection was evaluated on the Franka FR\,$3$. A trial was labeled successful if the part fully seated without jamming, retraction, or visible damage.

\begin{table}[h]
  \centering
  \caption{SnapNet detection rates on FR\,$3$.}
  \label{tab:results_counts}
  \begin{tabular}{ll}
    \toprule
    \textbf{Snap-fit Part}        & \textbf{Detection Rate} \\
    \midrule
    Marker Pen cap                & 14/15 (93.3\%) \\
    Highlighter cap               & 15/15 (100.0\%) \\
    Water Bottle Lid              & 15/15 (100.0\%) \\
    Type-C USB Cable                  & 13/15 (86.7\%) \\
    3D-printed Lens \& Frame      & 15/15 (100.0\%) \\
    E-stop                        & 15/15 (100.0\%) \\
    \midrule
    \textbf{Overall recall}       & \textbf{96.7\%} \\
    \bottomrule
  \end{tabular}
\end{table}

SnapNet achieved consistently high detection accuracy on the FR3 across a diverse set of snap-fit tasks. The only notable misses occurred with the Type-C USB Cable, where rapid but low-amplitude signatures occasionally evaded detection. Across all trials, inference latency remained below $50$\,ms, enabling reliable triggering of downstream impedance control.

\subsubsection{Bimanual Coordination via DS}
\label{subsec:res_ds}
We executed 15 pick-transport-insert trials with the DS policy. Tracking quality during coupled (\(\kappa=1\)) and decoupled (\(\kappa=0\)) phases was summarized by mean task-space error (mean$\,\pm\,$std over trials).

\begin{table}[h]
  \centering
  \caption{DS coordination metrics (mean tracking error).}
  \label{tab:ds_metrics}
  \begin{tabular}{lcc}
    \toprule
       & \(\epsilon_{\text{sync}}\) (mm) & \(\epsilon_{\text{dec}}\) (mm) \\
    \midrule
    Phases $1$-$2$ (coupled)                & \(2.1 \pm 0.9\)               & --- \\
    Phase $3$ (decoupled)                   & ---                           & \(2.6 \pm 1.1\) \\
    \bottomrule
  \end{tabular}
\end{table}

Errors remained within a few millimeters in both regimes, indicating that the phase-gated coupling preserves coordination during transport and allows precise unilateral insertion without controller switching.

\begin{figure}[ht]
  \centering
  \begin{subfigure}{0.8\linewidth}
    \centering
    \includegraphics[width=\textwidth]{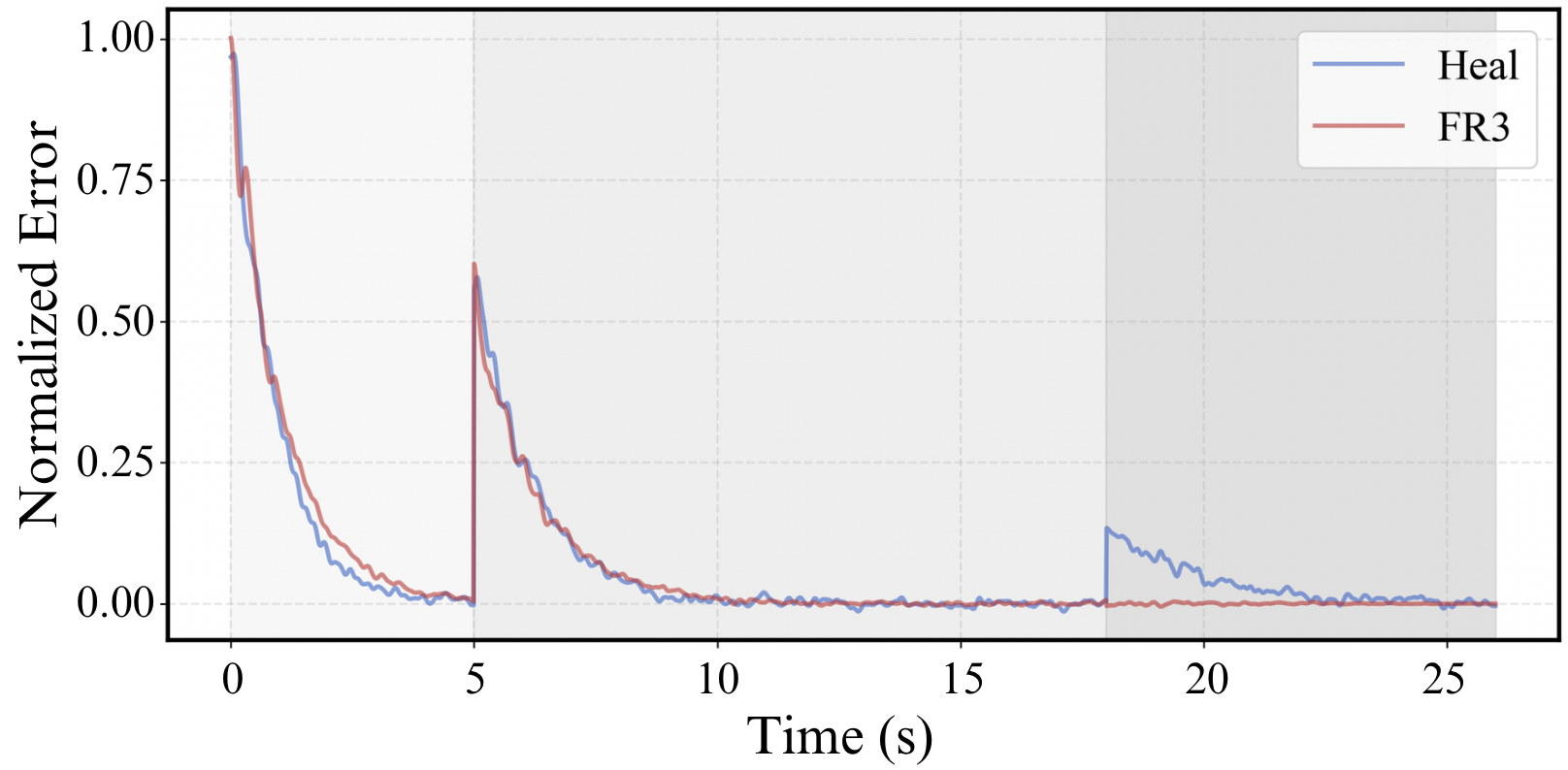}
    \caption{Normalized tracking error \(\epsilon_i(t)\).}
    \label{fig:ds_res_err}
  \end{subfigure}
  
  \vspace{1em}
  
  \begin{subfigure}{0.8\linewidth}
    \centering
    \includegraphics[width=\textwidth]{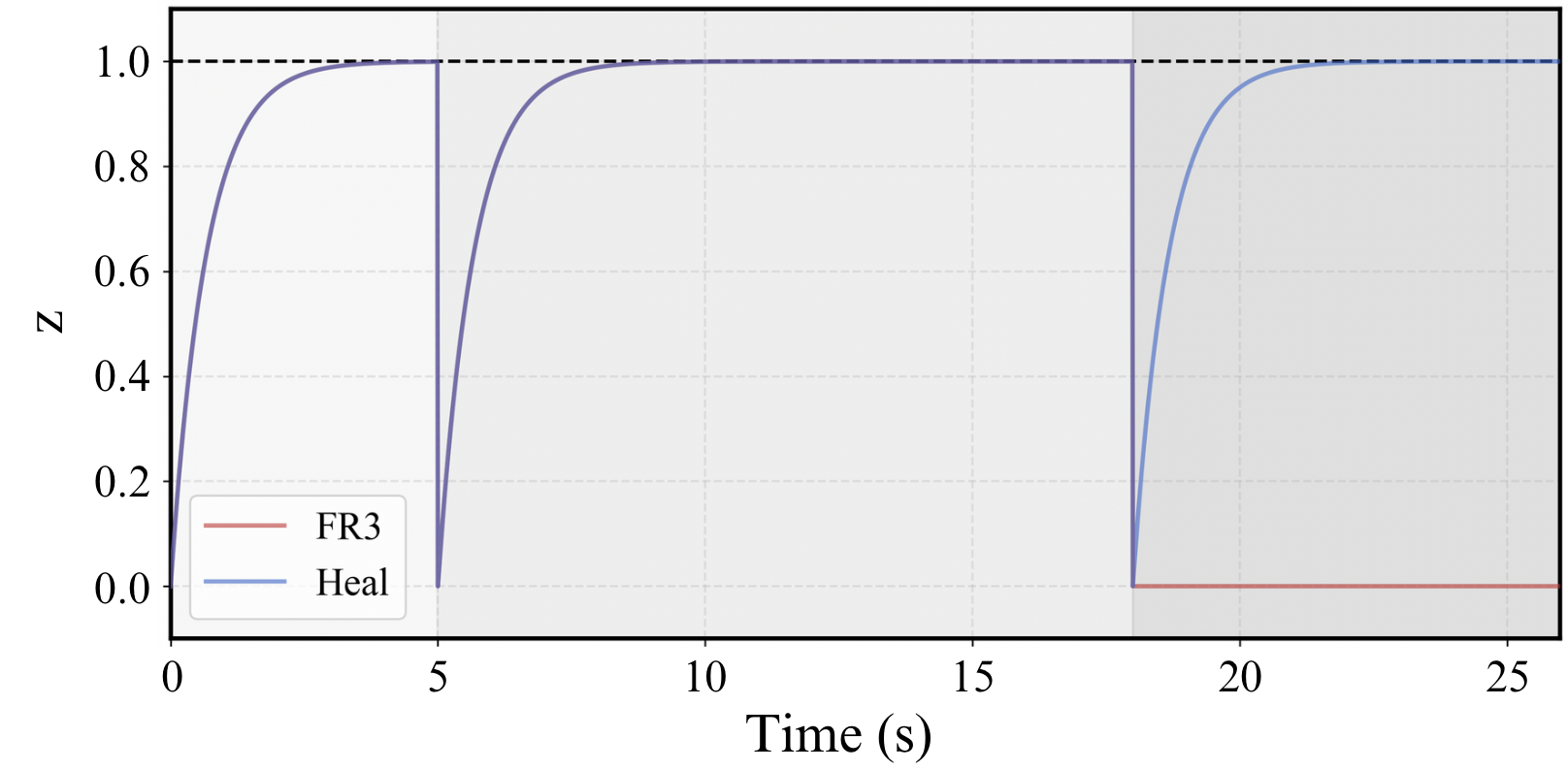}
    \caption{Phase progress \(z_i(t)\).}
    \label{fig:ds_res_phase}
  \end{subfigure}
  \caption{DS-based bimanual coordination results on Franka FR3 (red) and Addverb 
  Heal Cobot (blue). Shaded spans indicate Phase~1 (light gray), Phase~2 (medium gray), 
  and Phase~3 (dark gray).}
  \label{fig:ds_res}
\end{figure}

\subsubsection{Overall Framework}
\label{subsec:res_ablation}
We compared three insertion controllers on the lens-frame task, holding the coordination policy and trajectories fixed. This task due to its high force sensitivity and high overshoot propensity qualifies as delicate assembly task. Position control and constant-gain impedance represent standard baselines; the proposed VIC was triggered by SnapNet upon engagement.

\begin{table}[h]
  \centering
  \caption{Ablation on insertion scheme (lens-frame, $15$ trials).}
  \label{tab:assembly_ablation}
  \begin{tabular}{lcc}
    \toprule
    \textbf{Scheme}                    & \textbf{Success rate} & $\mathbf{F}_{\max}$ (N) \\
    \midrule
    Position control                   & \(6/15\) (40\%)       & \(25.1 \pm 3.8\)        \\
    Fixed Cartesian impedance          & \(11/15\) (73\%)      & \(22.4 \pm 4.2\)        \\
    Event-triggered VIC (proposed)     & \(15/15\) (100\%)     & \(16.3 \pm 3.1\)        \\
    \bottomrule
  \end{tabular}
\end{table}

Position control was sensitive to sub-millimeter misalignments, often leading to jamming. Constant impedance improved tolerance but did not attenuate the snap-induced impulse. Event-triggered VIC achieved perfect insertion on this task while reducing peak contact force by $\approx 30\%$ relative to constant impedance, indicating that axis-selective stiffness decay at the detected snap mitigates impact without sacrificing pre-engagement precision.
\section{DISCUSSION}
\label{sec:discussion}

The results in \S\ref{subsec:results} confirm three central claims. SnapNet provides reliable proprioceptive snap detection on the FR3, with high recall and $50$\,ms end-to-end latency. The DS-based coordinator maintains low tracking error in both coupled and decoupled phases and supports smooth transitions without controller switching. Event-triggered VIC reduces peak impact forces and prevents snap-through failures in the delicate assembly tasks, which surpass the performance of position control and fixed impedance as reported in Table~\ref{tab:assembly_ablation}. Together, these components provide brief-spike snap detection, shared progress tracking with controlled decoupling, and adaptive stiffness reduction along the insertion axis while preserving lateral alignment. This integrated structure yields high pre-engagement accuracy and rapid post-engagement energy dissipation, consistent with the mechanics described in \S\ref{subsubsec:mechanics}.

\begin{table}[h]
  \centering
  \caption{SnapNet detection on Addverb Heal cobot}
  \label{tab:robust_heal}
  \begin{tabular}{lc}
    \toprule
    \textbf{Snap-fit Part} & \textbf{Detection Rate} \\
    \midrule
    Marker Pen Cap               & 13/15 (86.7\%) \\
    Highlighter Cap              & 14/15 (93.3\%) \\
    Type-C USB Cable             &  8/15 (53.3\%) \\
    3D-printed Lens \& Frame     & 12/15 (80.0\%) \\
    E-stop                       & 15/15 (100.0\%) \\
    \midrule
    \textbf{Overall recall}      & \textbf{82.6\%} \\
    \bottomrule
  \end{tabular}
\end{table}

Joint velocity offered the strongest trade-off between reliability and practicality among all modalities benchmarked offline. It is always available, requires no additional hardware, and reflects the short impulse at engagement. Force and acceleration signals were also informative but require instrumentation. Contact microphones and tactile sensors produced inconsistent responses because their signatures depended on mounting and contact location.

SnapNet was also evaluated on the Heal platform, which has higher friction and lower passive compliance than the FR3. The velocity transients were smaller but often remained detectable, as shown in Table~\ref{tab:robust_heal}. This suggests that snap engagements generate platform-independent kinematic cues even when their amplitude is reduced.

The comparison on the lens-frame assemblies illustrates the need for precise motion before the snap and compliance after the snap. Position control offers accurate alignment but cannot handle the impact at engagement. Fixed impedance improves tolerance but still transmits a strong impulse. Event-triggered VIC reduces insertion-axis stiffness once SnapNet detects snap engagement and achieves a reduction of about thirty percent in peak force along with perfect success rates.

Although the framework performed well, several factors affect real-world deployment:

\begin{enumerate}[label=\arabic*.]
    \item \textbf{Platform back-drivability.} Robots with low back-drivability or high friction weaken the velocity transients that SnapNet relies on, which reduces detection quality.
    \item \textbf{Aggressive motion profiles.} Large accelerations or abrupt command changes can create velocity spikes that resemble snap events. Restricting detection to the final insertion phase and smoothing trajectories reduces this risk.
    \item \textbf{Workspace and configuration effects.} Some joint configurations attenuate snap signatures. Confining detection to a well-defined mating region improves consistency.
\end{enumerate}

This study focused on single-step insertions with one main engagement. Parts with multiple latches may require multiple detections and staged impedance updates. Industrial safety limits may also restrict allowable stiffness levels and decay rates. These factors provide opportunities for expanding the approach to a wider range of snap-fit assemblies.

\section{CONCLUSION}
\label{sec:conclusion}
This work introduced a unified framework for bimanual snap-fit assembly that combines fast proprioceptive engagement detection using SnapNet, phase-gated DS coordination, and event-triggered variable impedance control. On the FR3, SnapNet achieved high recall with low latency, the DS policy maintained millimeter-level tracking in both coupled and decoupled phases, and SnapNet-triggered VIC removed snap-through failures while lowering peak impact forces by approximately $30\,\%$ compared to position control and fixed impedance. Tests on a non-backdrivable platform showed that proprioceptive cues remain usable even when attenuated. 

The separation of detection and control, together with a compliance law that reduces insertion-axis stiffness after engagement while preserving lateral alignment, provides both accurate pre-snap positioning and effective post-snap energy absorption. Future work will focus on integrating the remaining elements of the assembly pipeline to move toward a fully end-to-end dual-arm snap-fit solution.

% This work presented a unified, deployment-oriented framework for bimanual snap-fit assembly that couples fast proprioceptive engagement detection (SnapNet), phase-gated DS coordination, and event-triggered variable impedance control. On the FR\,$3$, SnapNet detected snaps with high recall and low latency; the DS policy achieved millimeter-level tracking in both coupled and decoupled phases; and SnapNet-triggered VIC eliminated snap-through failures while reducing peak impact forces by $\approx30$\% on a low-tolerance task, outperforming position control and constant-gain impedance. A robustness check on a non-backdrivable platform confirmed that proprioceptive signatures remain informative, albeit attenuated. 

% By separating detection from control and embedding physical structure into the compliance law (axis-selective stiffness decay with passivity-preserving damping), the framework consolidates precise pre-engagement alignment with robust post-engagement energy absorption. Future work will aim to integrate the remaining components of the pipeline to realize a fully end-to-end framework for coordinated dual-arm snap-fit assembly. 

\appendices
\section{Proof of Theorem~1}
\label{app:thm1}

Consider the phase dynamics
\begin{equation}
\dot z_i = g_i(z_i) + \kappa\, g_c(z_i),
\end{equation}
where
\begin{align}
g_i(z_i) &= -k_1\,\frac{z_i - z_{d_i}}{(|z_i - z_{d_i}|+\varepsilon)^{1-\eta}}, \\
g_c(z_i) &= -k_2\,\frac{z_i - z_c}{(|z_i - z_c|+\varepsilon)^{1-\eta}},
\end{align}
with constants $k_1,k_2>0$, $\varepsilon>0$, and $\eta\in(0,1)$.

\paragraph{ Decoupled case $\kappa=0$}
Each arm evolves as
\begin{equation}
\dot z_i = -k_1\,\frac{z_i - z_{d_i}}{(|z_i - z_{d_i}|+\varepsilon)^{1-\eta}}.
\end{equation}
Let $e_i = z_i - z_{d_i}$ and choose
\begin{equation}
V_i(e_i)=\frac{1}{\eta}\!\left[(|e_i|+\varepsilon)^{\eta}-\varepsilon^{\eta}\right].
\end{equation}
Then
\begin{equation}
\dot V_i = -k_1\,|e_i|(|e_i|+\varepsilon)^{2\eta-2}\le0,
\end{equation}
with equality only at $e_i=0$. Hence $e_i(t)\!\to\!0$ globally, implying $z_i(t)\!\to\!z_{d_i}$. 

\paragraph{ Coupled case $\kappa=1$ and $z_{d_i}\equiv z_d$}
Let $e_i=z_i-z_d$ and define the mean $z_c=\tfrac{1}{N}\sum_j z_j$.  
Using the same $V_i$ as before, the composite Lyapunov function
\begin{equation}
V = \sum_i V_i + \gamma\,W, \qquad
W = \tfrac{1}{2}\sum_i (z_i-z_c)^2,
\end{equation}
with a small $\gamma>0$, yields
\begin{equation}
\dot V \le -c_1\!\sum_i |e_i|\phi_1(e_i) - c_2\!\sum_i (z_i-z_c)^2\phi_2(z_i-z_c),
\end{equation}
for some positive continuous $\phi_1,\phi_2$. Hence $\dot V\!\le\!0$, 
and the largest invariant set satisfying $\dot V=0$ is 
$\{z_i=z_c=z_d,\,\forall i\}$.  
Thus $z_i(t)\!\to\!z_d$ and $z_i-z_c\!\to\!0$ as $t\!\to\!\infty$.
\qed

\section{Proof of Theorem~2}
\label{app:thm2}

Let $e_i = p_i - \gamma_i(z_i)$ and $E_i=\|e_i\|$.  
Under the control law
\[
\dot p_i = -\alpha_i(E_i)e_i, \qquad 
\alpha_i(E_i)=\alpha_{\min} + 
(\alpha_{\max}-\alpha_{\min})\frac{E_i}{E_i+\delta},
\]
the closed-loop error dynamics are
\[
\dot e_i = -\alpha_i(E_i)e_i - J_{\gamma_i}(z_i)\dot z_i,
\]
where $\|J_{\gamma_i}(z_i)\|\le L_i$ (Lipschitz $\gamma_i$).  
With $V_i=\tfrac{1}{2}\|e_i\|^2$,
\begin{align}
\dot V_i &= -\alpha_i(E_i)\|e_i\|^2 - e_i^\top J_{\gamma_i}(z_i)\dot z_i
  \le -\alpha_{\min}\|e_i\|^2 + L_i\|e_i\||\dot z_i|.
\end{align}
Applying Young’s inequality,
\[
L_i\|e_i\||\dot z_i| 
\le \frac{\alpha_{\min}}{2}\|e_i\|^2 
   + \frac{L_i^2}{2\alpha_{\min}}|\dot z_i|^2,
\]
and therefore
\begin{equation}
\dot V_i \le -\tfrac{\alpha_{\min}}{2}\|e_i\|^2
             + \tfrac{L_i^2}{2\alpha_{\min}}|\dot z_i|^2.
\end{equation}
Integrating yields
\[
\|e_i(t)\|^2 
\le \|e_i(0)\|^2 e^{-\alpha_{\min}t}
   + \frac{L_i^2}{\alpha_{\min}^2}\sup_{s\le t}|\dot z_i(s)|^2.
\]
As $z_i(t)\!\to\!z_{d_i}$ and $\dot z_i(t)\!\to\!0$, 
the second term vanishes and $e_i(t)\!\to\!0$, and hence 
$p_i(t)\!\to\!\gamma_i(z_{d_i})$.
\qed

\bibliographystyle{IEEEtran}
\bibliography{ref}

\end{document}